\documentclass[graybox]{svmult}

\usepackage{mathptmx}       
\usepackage{helvet}         
\usepackage{courier}        
\usepackage{type1cm}        
\usepackage{makeidx}         
\usepackage{graphicx}        
\usepackage{multicol}        
\usepackage[bottom]{footmisc}
\usepackage{url}
\usepackage{multirow}

\usepackage{epstopdf}
\usepackage{multirow}
\usepackage{booktabs}
\usepackage{todonotes}
\usepackage{lipsum}
\usepackage{hyperref}
\hypersetup{hidelinks}

\usepackage[misc]{ifsym}
\usepackage{ifthen}
\usepackage{makeglos}

\graphicspath{{./}{./graphics/}{./graphics/eps/}{./graphics/pdf/}}

\usepackage{multirow}
\usepackage[official]{eurosym}
\usepackage{subfigure}

\newcommand{\buildbook}{false}

\makeindex             

\makeglossary

\begin{document}

%
%
%

\ifthenelse{\equal{true}{\buildbook}}{
\title{Review of Face Presentation Attack Detection Competitions}
}
{
\title*{Chapter 28\\
Face Presentation Attack Detection}
}

\def\ie{{\em i.e.}}
\def\eg{{\em e.g.}}
\def\etal{{\em et al.}}
\def\vs{\emph{vs.}}


\author{Zitong Yu, Chenxu Zhao, and Zhen Lei \Letter}


\institute{Zitong Yu \at ROSE Lab, Nanyang Technological University, Singapore 639798, Singapore\\ \email{zitong.yu@ntu.edu.sg}
\and Chenxu Zhao \at SailYond Technology, Beijing 100083, China\\ \email{zhaochenxu@sailyond.com}
\and Zhen Lei \quad  \at National Laboratory of Pattern Recognition, Institute of Automation, Chinese Academy of Sciences, Beijing 100190, China\\
School of Artificial Intelligence, University of Chinese Academy of
Sciences (UCAS), Beijing 100049, China\\
Centre for
Artificial Intelligence and Robotics, Hong Kong Institute of Science \&
Innovation, Chinese Academy of Sciences, Hong Kong, China\\
\email{zlei@nlpr.ia.ac.cn}  (Corresponding author: Zhen Lei)}

%

%
\maketitle

\section{ }

\subsection{Introduction}
\label{sec:introduction}
Face recognition technology has been widely used in daily interactive applications such as checking-in and mobile payment due to its convenience and high accuracy. However, its vulnerability to presentation attacks (PAs) limits its reliable use in ultra-secure applicational scenarios. A presentation attack is first defined in ISO standard~\cite{ACER} as: a presentation to the biometric data capture subsystem with the goal of interfering with the operation of the biometric system. Specifically, PAs range from simple 2D print, replay and more sophisticated 3D masks and
partial masks. To defend the face recognition systems against PAs, both academia and industry have paid extensive attention to developing face presentation attack detection (PAD)~\cite{yu2021deep} technology (or namely `face anti-spoofing (FAS)').

In the early stage, commercial PAD systems are usually designed based on strong prior knowledge of obvious and macro liveness cues such as eye-blinking~\cite{li2008eye}, face and head movement~\cite{wang2009face} (e.g., nodding and smiling), gaze tracking~\cite{ali2012liveness}. They assume that the 2D print attacks are static and lack of \textit{interactive} dynamic cues. Despite easy development and deployment, these methods suffer from high false acceptance errors when replayed face videos or partial wearable 3D masks are presented to mimic the interactive liveness cues. To eliminate the requirement of interactive dynamics and explore more intrinsic and micro features for face PAD, plenty of traditional handcrafted feature~\cite{li2016generalized,Komulainen2014Context,Boulkenafet2016Face} based methods are then proposed for face PAD. On the one hand, according to the evidence that PAs have degraded static/dynamic texture details and spoof artifacts (e.g., moiré pattern), classical handcrafted texture descriptors (e.g., LBP~\cite{Boulkenafet2016Face} and HOG~\cite{Komulainen2014Context}), image quality assessment~\cite{galbally2013image} metrics and micro motion~\cite{tirunagari2015detection} features are designed for extracting effective spoofing patterns from various color spaces (e.g., RGB, HSV, and YCbCr). On the other hand, considering the facts that 3D mask attacks might contain realistic textural appearance but without quasi-periodic live physiological cues, facial video-based remote physiological signals (e.g., rPPG~\cite{yu2022benchmarking,yu2021transrppg}) measurement technique is introduced for 3D high-fidelity mask attack detection.


\newcommand{\tabincell}[2]{\begin{tabular}{@{}#1@{}}#2\end{tabular}}

Subsequently, with the development of deep learning for computer vision tasks and release of larger-scale and diverse face PAD datasets~\cite{zhang2020celeba,liu2019deep,george2019biometric}, plenty of end-to-end deep learning based methods~\cite{yu2020searching,yu2020face,Liu2018Learning,yang2019face,Atoum2018Face,yu2020multi,zhang2020casia} are proposed for face PAD. Similar to many binary classification tasks (e.g., gender classification), many works~\cite{yang2014learn,Li2017An,george2019deep,jourabloo2018face,jia20203d,li2020compactnet} treat face PAD as a binary bonafide/PA classification problem thus utilizing a simple binary cross-entropy loss for model supervision. Recently, researchers find that binary cross-entropy loss cannot provide explicit task-aware supervision signals thus the models supervised by such loss easily learn unfaithful patterns~\cite{Liu2018Learning} (e.g., bezel and background). To alleviate this issue, more and more recent works focus on leveraging auxiliary pixel-wise supervision~\cite{Atoum2018Face,Liu2018Learning,yu2020face,george2019deep,yu2020fas2} to provides more fine-grained context-aware supervision signals. For example, according to the geometric discrepancy between bonafide with facial depth and flat 2D PAs, pseudo depth labels~\cite{Atoum2018Face,Liu2018Learning} are designed to force model learning local geometric cues for bonafide/PA discrimination. Similarly, pixel-wise auxiliary supervisions with reflection maps~\cite{yu2020face} and binary mask label~\cite{george2019deep,liu2019deep} benefit models to describe the local physical cues in pixel/patch level.

In real-world scenarios, different dominant conditions (e.g., illumination, face resolution and sensor noise)
influence the face PAD system a lot. For instance, a well-designed and -trained deep PAD model on normal illumination and high face resolutions might perform poorly under low-light and low-resolution scenario due to their large distribution gaps. Thus, to learn more generalized and robust features against unseen domain shifts is vital for practical applications. To this end, domain adaptation~\cite{li2018unsupervised} and generalization~\cite{shao2019multi,wang2022domain,jia2020single} techniques are introduced to enhancing the generalization capacity of the deep face PAD models. The former leverages the knowledge from target domain to bridge the gap between source and target domains while the latter helps PAD models learn generalized feature representation from multiple source domains directly without any access to target data.

The structure of this chapter is as follows. Section~\ref{sec:background} introduces the research background, including presentation attacks, pipeline for face PAD, and existing datasets. Section~\ref{sec:method} reviews the handcrafted feature based and deep learning based methods for face PAD. Section~\ref{sec:experiment} provides experimental results and analysis. Section~\ref{sec:Applications} discusses the practical face PAD based industrial applications. Finally, Section~\ref{sec:conclusion} summarizes the conclusions and lists some future challenges.

\subsection{Background}
\label{sec:background}

Here, we will introduce the common face PAs and general face PAD pipeline in face recognition system first. Then, mainstream camera sensors for face PAD are presented. Finally, existing face PAD datasets are summarized.

\subsubsection{Face Presentation Attacks}

\begin{figure} [t]
\centering
\includegraphics[width=1.0\linewidth]{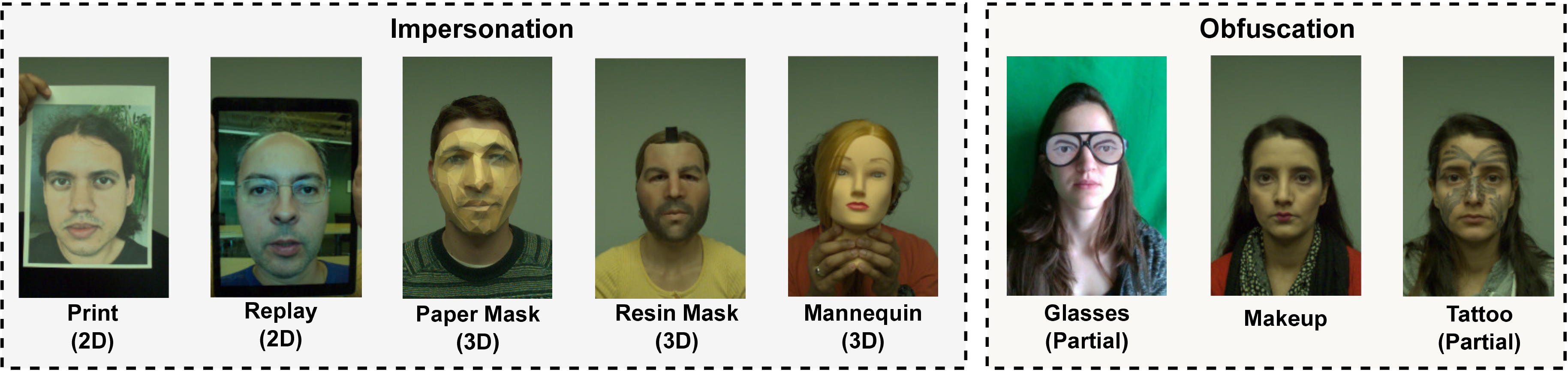}
  \caption{ 
  Visualization of different types of face presentation attacks~\cite{heusch2020deep}.
  }
\label{fig:dataset}
\end{figure}

Face presentation attacks usually mislead the real-world automatic face recognition (AFR) systems via presenting face upon physical mediums (e.g., photograph, a video, or a 3D mask) of a targeted or obfuscated person in front of the imaging sensors. Some representative PA types are illustrated in Fig.~\ref{fig:dataset}. According to the intention that whether the attackers would mimic targeted identities or hidden their own identities, face PAs~\cite{marcel2019handbook} can be divided into two categories: 1) \textit{impersonation}, which spoofs the AFR systems recognized as someone else via copying a genuine user’s facial attributes to presentation instruments such as photo, electronic screen, and 3D mask; and 2) \textit{obfuscation}, which decorates the face to hide or remove the attacker’s own identity via wearing glasses/wig or with makeup/tattoo.

PAs are broadly classified into \textit{2D} and \textit{3D} attacks according to the geometric depth. Common 2D PAs usually contain print and replay attacks such as flat/wrapped printed photos, eye/mouth-cut photos, and replay of face videos on electronic screens. Compared with traditional 2D PAs, 3D presentation attacks such as 3D face masks and mannequins are more realistic in terms of color, texture, and geometry structure, which can be made of different materials, e.g., paper, resin, plaster, plastic, silicon and latex. According to the proportion of facial region covering, PAs can be also categorized to \textit{whole} or \textit{partial} attacks. Compared with common PAs covering the whole face, a few partial attacks only presented on partial facial regions. For example, attackers would cutout the eye regions from the print face photo to spoof the eye blinking based PAD system while funny eyeglasses with adversarial patterns would be worn in the eyes region to attack the face PAD algorithms. Compared with attacks on whole face, partial attacks are more obscure and challenging to defend.  



\begin{figure}[t]
\centering
\includegraphics[width=0.95\linewidth]{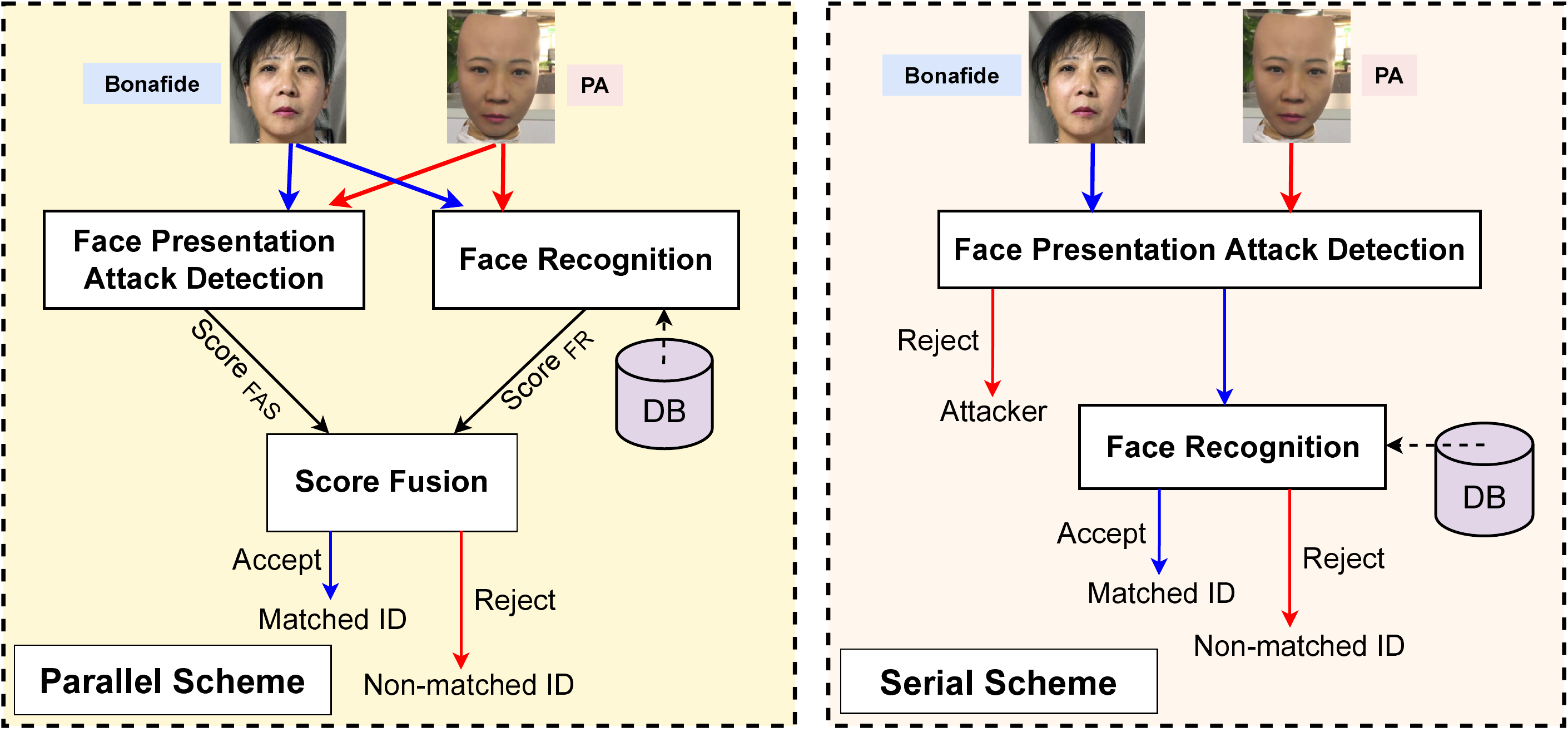}
  \caption{ 
  Typical face PAD pipeline. Face PAD could be integrated with face recognition systems with (left) parallel or (right) serial scheme for reliable face ID matching.
  }
\label{fig:pipeline}
\end{figure}

\subsubsection{Face PAD Pipeline in Face Recognition Systems}
The pipeline of face PAD in automatic face recognition systems (AFR) is illustrated in Fig.~\ref{fig:pipeline}. There are \textit{parallel} and \textit{serial} schemes for integrating face PAD with AFR systems. As for the \textit{parallel} scheme, the detected faces can be passed over the face PAD and AFR modules to obtain the respective predicted scores, which are then used for parallel fusion. The combined new final score is used to determine whether the sample comes from a genuine user or not. The parallel scheme is suitable to deploy in multi-core or multi-thread systems with good parallel computation specifications to perform PAD and face recognition simultaneously. Besides, the parallel scheme leaves the space for robust fusion strategies design with PAD and AFR outputs. In terms of \textit{serial} scheme, detected faces are first forwarded to PAD module to reject the PAs, and only the filtered bonafide faces can go through the face recognition phase. Despite delayed PAD time for the subsequent AFR of bonafide access attempts, the serial scheme avoids extra work to the AFR module in the case of PAs, since the PAs have been detected in an early stage.

\subsubsection{Camera Sensors for Face PAD}

\begin{table}[t]
\centering
\caption{Comparison with camera sensors for face PAD under 2 environments (lighting condition and distance) and 3 PA types (print, replay and 3D mask). `NIR', `TOF', `SL' are short for `Near Infrared', `Time of Flight', `Structured Light', respectively.}


\resizebox{0.95\textwidth}{!}{
\begin{tabular}{|c|c|c|c|c|c|c|}
\hline
\multirow{2}{*}{\textbf{\quad Sensor \quad}} & \multirow{2}{*}{\textbf{\quad Cost \quad}} &\multicolumn{2}{c|}{\textbf{Environment}} &\multicolumn{3}{c|}{\textbf{Attack Type}}\\
\cline{3-7} 
& & \tabincell{c}{\quad Lighting \quad} &\tabincell{c}{\quad Distance \quad} &\tabincell{c}{ \quad Print \quad} &\tabincell{c}{ \quad Replay \quad}&\tabincell{c}{\quad 3D Mask \quad}\\
\hline
RGB & Cheap & Poor & Good & Medium & Medium & Medium \\
\hline
Depth (TOF) & Medium & Good & Good & Good & Good & Poor \\
\hline
Depth (SL) & Cheap & Medium & Poor & Good & Good & Poor \\
\hline
NIR & Cheap & Good & Poor & Good & Good & Medium \\
\hline
RGB-NIR & Medium & Good & Medium & Good & Good & Good \\
\hline
RGB-Depth & Medium & Medium & Medium & Good & Good & Medium \\
\hline
RGB-NIR-Depth & Expensive & Good & Good & Good & Good & Good \\
\hline
\end{tabular}
}
\label{tab:sensorstemp}
\end{table}


Commercial RGB camera-based face PAD has been widely used in daily face recognition applications like mobile unlocking due to its acceptable security and low hardware cost. Besides visible RGB modality, depth and near infrared (NIR) modalities are also widely used in practical PAD deployment with acceptable costs. As for the depth camera, accurate 3D depth geometric surface of the captured face can be measured, which is very appropriate for flat 2D PAD without rich 3D facial cues. Two representative kinds of depth sensors are Time of Flight (TOF) and 3D Structured Light (SL). Compared with SL, TOF is more robust to environmental conditions such as distance and outdoor lighting but more expensive. Depth cameras with TOF or SL are usually embedded in mainstream mobile phone platforms (e.g., Iphone, Sumsung, OPPO, and Huawei) to benefit the RGB-Depth based face PAD. In terms of NIR cameras~\cite{sun2016context}, they contain complementary spectrums besides RGB, which explicitly capture material-aware reflection discrepancy between bonafide faces and PAs but are sensitive to long distance. In addition, RGB-NIR integration hardware modules are also popular in access control systems due to their high performance-price ratio. In real-world deployment with high-security needs, integrating with all three modalities (RGB-NIR-Depth) usually provides the most robust performance in terms of environmental conditions (lighting and distance) and attack types (print, replay, and 3D mask). The characteristics of different sensors for face PAD are compared in Table~\ref{tab:sensorstemp}. Visualization of typical bonafide and PA samples with RGB, Depth, and NIR modalities are illustrated in Fig.~\ref{fig:Multi-modalities}.  


\subsubsection{Face PAD Datasets}

\begin{table*}
\centering
\caption{A summary of unimodal face PAD datasets. `\#Sub.', `I' and `V' are short for `Subjects', `images' and `videos', respectively.} \label{tab:dataset}
\resizebox{1.0\textwidth}{!} {\begin{tabular}{c c c c c} 
 \toprule[1pt]
 Dataset \& Reference & Year & \#Bonafide/PA & \#Sub.  & Attack Types \\
 \midrule
 NUAA~\cite{tan2010face} & 2010 & 5105/7509(I) & 15   &  Print(flat, wrapped)\\

 \midrule
 YALE\_Recaptured~\cite{peixoto2011face} & 2011 & 640/1920(I) & 10  &  Print(flat)\\

 \midrule
 CASIA-MFSD
~\cite{Zhang2012A} & 2012 & 150/450(V) & 50  & Print(flat, wrapped, cut), Replay(tablet)\\

 \midrule
REPLAY-ATTACK
~\cite{ReplayAttack} & 2012 & 200/1000(V) & 50 & Print(flat), Replay(tablet, phone)\\

 \midrule
MSU-MFSD
~\cite{wen2015face} & 2014 & 70/210(V) & 35  & Print(flat), Replay(tablet, phone)\\

 \midrule
REPLAY-Mobile
~\cite{costa2016replay} & 2016 & 390/640(V) & 40 & Print(flat), Replay(monitor)\\

 \midrule
HKBU-MARs V2
~\cite{liu20163d} & 2016 & 504/504(V) & 12 & \tabincell{c}{Mask(hard resin) from \\ Thatsmyface and REAL-f}\\

 \midrule
MSU USSA
~\cite{Patel2016Secure} & 2016 & 1140/9120(I) & 1140 & Print(flat), Replay(laptop, tablet, phone)\\


 \midrule
OULU-NPU
~\cite{Boulkenafet2017OULU} & 2017 &  720/2880(V)  & 55  & Print(flat), Replay(phone)\\

 \midrule
Rose-Youtu
~\cite{li2018unsupervised} & 2018 &  500/2850(V) & 20   & \tabincell{c}{Print(flat), Replay(monitor, laptop), \\Mask(paper, crop-paper)}\\

 \midrule
SiW
~\cite{Liu2018Learning} & 2018 &  1320/3300(V)  & 165  & \tabincell{c}{Print(flat, wrapped), \\Replay(phone, tablet, monitor)}\\

 \midrule
WFFD
~\cite{jia20203d} & 2019 & \tabincell{c}{2300/2300(I)\\140/145(V)} & 745  & Waxworks(wax)\\

 \midrule
SiW-M
~\cite{liu2019deep} & 2019 & 660/968(V) & 493  & \tabincell{c}{Print(flat), Replay, Mask(hard resin, \\plastic, silicone, paper, Mannequin),\\ Makeup(cosmetics, impersonation, \\Obfuscation), Partial(glasses, cut paper)}\\

 \midrule
Swax
~\cite{vareto2020swax} & 2020 &  \tabincell{c}{Total 1812(I)\\110(V) } & 55 & Waxworks(wax)\\

 \midrule
CelebA-Spoof
~\cite{zhang2020celeba} & 2020 &  \tabincell{c}{156384/\\469153(I)} & 10177  & \tabincell{c}{Print(flat, wrapped), Replay(monitor, \\tablet, phone), Mask(paper)}\\

 \midrule
\tabincell{c}{CASIA-SURF\\3DMask}
~\cite{yu2020fas2} & 2020 &  288/864(V) & 48  & Mask(mannequin with 3D print)\\

 \midrule
\tabincell{c}{HiFiMask}
~\cite{liu2021contrastive} & 2021 &  13650/40950(V) & 75  & Mask(transparent, plaster, resin)\\

 \bottomrule[1pt]
 \end{tabular}}
\end{table*}

\begin{table*}
\centering
\caption{A summary of multimodal face PAD datasets. `SWIR' is short for short-wave infrared.} \label{tab:multimodaldataset}
\resizebox{1.0\textwidth}{!} {\begin{tabular}{c c c c c c} 
 \toprule[1pt]
 Dataset \& Reference & Year & \#Bonafide/PA & \#Sub. & Sensor & Attack Types \\

 \midrule[1pt]
 
3DMAD
~\cite{erdogmus2014spoofing} & 2013 & 170/85(V) & 17 & RGB, Depth  & Mask(paper, hard resin)\\

 \midrule
MLFP
~\cite{agarwal2017face} & 2017 & 150/1200(V) & 10 & \tabincell{c}{RGB, NIR, \\Thermal}  & Mask(latex, paper)\\

 \midrule
CSMAD
~\cite{bhattacharjee2018spoofing} & 2018 &  104/159(V+I)  & 14 & \tabincell{c}{RGB, Depth, \\NIR, Thermal}  & Mask(custom silicone)\\

 \midrule
CASIA-SURF
~\cite{casiasurf} & 2019 &  \tabincell{c}{3000/\\18000(V)} & 1000 &  \tabincell{c}{RGB, Depth,\\ NIR}  & Print(flat, wrapped, cut)\\

 \midrule
WMCA
~\cite{george2019biometric} & 2019 & 347/1332(V) & 72 &  \tabincell{c}{RGB, Depth,\\ NIR, Thermal}   & \tabincell{c}{Print(flat), Replay(tablet),\\ Partial(glasses), Mask(plastic, \\silicone, and paper, Mannequin)}\\

 \midrule
CeFA
~\cite{li2020casia} & 2020 &   \tabincell{c}{6300/\\27900(V)} & 1607 & \tabincell{c}{RGB, Depth,\\NIR}  & \tabincell{c}{Print(flat, wrapped), Replay, \\Mask(3D print, silica gel)}\\

 \midrule
HQ-WMCA
~\cite{heusch2020deep} & 2020 &  555/2349(V) & 51 & \tabincell{c}{RGB, Depth, \\NIR, SWIR,\\Thermal}  & \tabincell{c}{Laser or inkjet  Print(flat), \\Replay(tablet, phone), Mask(plastic, \\silicon, paper, mannequin), Makeup,\\ Partial(glasses, wigs, tatoo)}\\

 \midrule
PADISI-Face
~\cite{rostami2021detection} & 2021 & 1105/924(V) & 360 &  \tabincell{c}{RGB, Depth,\\ NIR, SWIR, \\Thermal}  & \tabincell{c}{Print(flat), Replay(tablet, phone),\\Partial(glasses,funny eye), Mask(plastic, \\silicone, transparent, Mannequin)}\\

 \bottomrule[1pt]
 \end{tabular}}
\end{table*}

In the past decade, a few face PAD datasets are established for training new PAD techniques and evaluating their performance against domain shifts and PAs types. Detailed statistics and descriptions of publicly available unimodal and multimodal face PAD datasets are summarized in Table~\ref{tab:dataset} and Table~\ref{tab:multimodaldataset}, respectively. 


In terms of RGB-based unimodal face PAD datasets shown in Table~\ref{tab:dataset}, there are only five public datasets~\cite{tan2010face,peixoto2011face,Zhang2012A,ReplayAttack,wen2015face} at the early stage from year 2010 to 2015. Due to the immature 3D mask manufacturing process with high cost at that time, these datasets only contain 2D PAs (i.e., print and replay attacks) and with limited subjects (no more than 50), which have insufficient data scale and attack diversity for generalized face PAD training and evaluation. Subsequently, there are two main trends for unimodal dataset development: 1) \textit{larger-scale subjects and data amount}. For example, the recently released datasets CelebA-Spoof~\cite{zhang2020celeba} and HiFiMask~\cite{liu2021contrastive} contain more than 600000 images and 50000 videos, respectively. Besides, MSU USSA~\cite{Patel2016Secure} and CelebA-Spoof~\cite{zhang2020celeba} are recorded with more than 1000 and 10000 subjects, respectively. 2) \textit{diverse attack types}. Besides common 2D print and replay attacks, more and more sophisticated 3D attacks and novel partial attacks are considered in recent face PAD datasets. For instance, there are high-fidelity 3D mask attacks made of different kinds of materials (e.g., 3D print, plaster, resin) in HKBU-MARs V2~\cite{liu20163d} and HiFiMask~\cite{liu2021contrastive}. As shown in shown in Table~\ref{tab:multimodaldataset}, similar trends of larger-scale subjects and data amount as well as attack types can be found in the development of multimodal face PAD datasets. Moreover, it can be observed that \textit{more kinds of modalities} are collected in recent face PAD datasets. For example, HQ-WMCA~\cite{heusch2020deep} and PADISI-Face~\cite{rostami2021detection} contain five modalities (RGB, Depth, NIR, short-wave infrared (SWIR), and Thermal).

\subsection{Methodology}
\label{sec:method}

To determine the liveness of user's faces during the identity verification procedure, \textit{interactive} face PAD methods are usually adopted. However, such interactive instructions (e.g., eye-blinking, facial expression, head movement, and vocal repeating) require the users' long-term participation, which are unfriendly and inconvenient. Thanks to the recent software-based methods designed with rich face PAD cues, the \textit{silent} face PAD system could automatically and quickly detect the PAs without any user interactions. In this subsection, we summarize the classical handcrafted PAD features and recent deep learning based methods for silent face PAD.

\subsubsection{Handcrafted Feature based Face PAD}
According to the features properties, we introduce the handcrafted feature based face PAD approaches based on five main cues, i.e., structural material, image quality, texture, micro motion, and physiological signals. The handcrafted features are usually cascaded with a support vector machine (SVM) or a multi-layer perception (MLP) for binary classification to distinguish bonafide faces from PAs. 

\vspace{0.4em}
\noindent\textbf{Structural Material based Approaches.}\quad   
In real-world cases, PAs are always broadcasted by physical presentation attack instruments (PAIs) (e.g., paper, glass screen and resin mask), which have obvious material properties difference with human facial skin. Such difference can be explicitly described as meaningful spoofing cues (e.g., structural depth and specular reflection) for face PAD. In order to obtain the 3D structure or material of the face, the most direct way is to use a binocular/depth or SWIR cameras. However, as a single RGB camera is the most common hardware configuration in practical applications, lots of face PAD research works still focus on 3D and material cue estimation based on the monocular RGB camera. On one hand, based on the assumption that 2D PAs on paper and screen are usually flat and without depth information, Wang et al.~\cite{wang2013face} propose to recover the sparse 3D shape of face images to detect various 2D attacks. On the other hand, the illumination and reflection discrepancy of the structural materials between human facial skin and 2D PAs are used as important spoof cues. Kim et al.~\cite{de2012moving} utilize the illumination diffusion cues based on the fact that illumination from 2D surfaces of 2D attacks diffuses slower and has a more uniform intensity distribution than 3D surfaces. Besides, Wen et al.~\cite{wen2015face} propose to calculate the statistical features based on the percentage of the specular reflection components from face image to detect the screen replay attacks. The methods based on the structural material cues have great rationality to detect the 2D PAs theoretically. However, estimating depth and material information from a monocular RGB camera is an ill-conditioned problem, and the computational complexity of these methods is high.

\vspace{0.4em}
\noindent\textbf{Image Quality based Approaches.}\quad   
As the spoof faces are usually broadcasted of the real face from specific physical PAIs, the corresponding face image quality might be degraded due to the color distortion and instrument artifacts, which can be utilized as a significant cue for face PAD. Galbally et al.~\cite{galbally2013image} adopt 25 (21 full-reference and 4 non-reference) image quality assessment (IQA) metrics for face liveness detection. Wen et al.~\cite{wen2015face} employ three kinds of different IQA features (blurriness, color moment and color difference) for face PAD, which can effectively represent the intrinsic distortion of spoof images. Image quality based methods are effective for screen-replayed faces, low-quality printed faces, and rough 3D mask spoof face detection. However, high-quality printed faces as well as high-fidelity 3D mask faces would result in high false acceptance rates for these methods.

\vspace{0.4em}
\noindent\textbf{Texture based Approaches.}\quad   
Due to the PAI properties, textural details in spoof faces is usually coarse and smoothed. In contrast, bonafide faces captured via cameras directly keep more fine-grained local texture cues. Based on this evidence, many texture based approaches have been developed for face PAD. Specifically, several classical local texture descriptors such as local binary pattern (LBP)~\cite{maatta2011face} and histogram of oriented gradients (HOG)~\cite{Komulainen2014Context} are used to capture fine-grained texture features from face images. Based on the observation that texture features in the HSV color space are more invariant across different environments, Boulkenafet et al.~\cite{Boulkenafet2016Face} propose to extract LBP based color texture features from HSV space, which is efficient and generalized. 
However, the texture based methods rely on high-resolution input to distinguish subtle texture differences between bonafide and spoofing faces. If the image quality is not good enough, it will result in a high false rejection rate. In addition, due to the diversity of image acquisition conditions and spoofing instruments, extracted texture patterns are also variant, which makes it generalize poorly under complex real-world scenarios.

\vspace{0.4em}
\noindent\textbf{Micro Motion based Approaches.}\quad   
Liveness detection by capturing the user's short-term micro motion characteristics without interaction is feasible as facial dynamics (e.g., expression and head movement) or dynamic textures from live and spoof samples are distinguishable. Tirunagari et al.~\cite{tirunagari2015detection} propose to temporally magnify the facial motion first, and then extract two kinds of dynamic features including the histogram of oriented optical flow (HOOF) and Local Binary Pattern histograms from Three Orthogonal Planes (LBP-TOP) for face PAD. However, motion magnification usually brings external noises, which influences the robustness of the subsequent feature representation. Instead of motion magnification, Siddiqui et al.~\cite{siddiqui2016face} employ dynamic mode decomposition to select the most reliable dynamic mode for facial motion feature extraction. However, the micro motion based methods are not effective for wrapped/shaking paper attack and video replay attacks due to interference of undesired dynamics. These methods assume that there is a clear non-rigid motion discrepancy between bonafide and PAs, but in fact such micro motion is quite difficult to describe and represent explicitly.

\vspace{0.4em}
\noindent\textbf{Remote Photoplethysmograph based Approaches.}\quad   
Physiological signal is another important living body signal, and it is also an intrinsic cue for distinguishing live faces from artificial materials. In recent years, remote photoplethysmograph (rPPG) technology~\cite{yu2021facial} has developed quickly, which aims at measuring blood pulse flow by modeling the subtle skin color changes caused by the heartbeat. Due to the low transmittance characteristics of artificial materials, rPPG signals from the live faces are usually periodic, but more noisy on the PAs such as 3D mask and printed paper. Therefore, rPPG signals are suitable for face liveness detection.
Li et al.~\cite{li2016generalized} analyze the live/spoof rPPG cues via calculating the statistics of the rPPG frequency responses. Different from the method of spectrum analysis using long-term observation of rPPG signals in the frequency domain, Liu et al.~\cite{liu2020temporal} propose to leverage the temporal similarity of facial rPPG signals for fast 3D mask attack detection, which can be  within one second by analyzing the time-domain waveform of the rPPG signal. However, rPPG cues are sensitive to the head motion, light condition, and video quality. Another disadvantage is that the replayed video attack on electronic screen might still contain weak periodic rPPG signals.

\subsubsection{Deep Learning based Face PAD}

With the data-driven deep learning fashion in computer vision, deep neural networks have also been widely used in face PAD. Here we highlight some traditional deep learning approaches with cross-entropy and pixel-wise supervision first, and then introduce domain generalized deep learning methods.


\vspace{0.4em}
\noindent\textbf{Traditional Deep Learning Approaches with Cross-Entropy Supervision.}\quad   
As face PAD can be intuitively treated as a binary (bonafide vs. PA) or multi-class (e.g., bonafide, print, replay, mask) classification task, numerous deep learning methods are directly supervised with cross-entropy (CE) loss. Given an extracted face input $X$, deep PAD features $F$ can be represented via forwarding the deep models $\Phi$, and then the cascaded classification heads make the binary predictions $Y$, which are supervised by the binary cross-entropy (BCE) loss
\begin{equation} 
L_{BCE}=-(Y_{gt}log(\Phi(X))+(1-Y_{gt})log(1-\Phi(X))),
\end{equation}
where $Y_{gt}$ is the ground truth ($Y_{gt}=0$ for PAs and $Y_{GT}=1$ for bonafide. Supervised with BCE loss, Yang et al.~\cite{yang2014learn} propose the first end-to-end deep face PAD method using shallow convolutional neural networks (CNN) for bonafide/PA feature representation. Through the stacked convolution layers, CNN is able to capture the semantic spoof cues (e.g., hand-hold contour of the printed paper). However, training CNN from scratch easily overfits in the face PAD task due to the limited data amount and coarse supervision signals from BCE loss. To alleviate these issues, on the one hand, some recent researches~\cite{chen2019attention,george2020effectiveness} usually finetune the ImageNet-pretrained models (e.g., ResNet18 and  vision transformer) with BCE loss for face PAD. Transferring the well-tuned model parameters on large-scale generic object classification task to downstream face PAD data is relatively easier and efficient. On the other hand, a few works modify BCE loss into multi-class CE version to provide CNNs more fine-grained and discriminative supervision signals. Xu et al.~\cite{xu2020improving} rephrase face PAD as a fine-grained classification problem and propose to supervise deep model with multi-class (e.g., bonafide, print, and replay) CE loss. In this way, the intrinsic properties from bonafide as well as particular attack types could be explicitly represented. However, models supervised with multi-class CE loss still suffer from unsatisfactory convergence due to the class imbalance. Another issue is that these supervision signals with only global constraints might cause face PAD models to easily overfit to unfaithful patterns but neglect vital local spoof patterns. 

\begin{figure}[t]
\centering
\includegraphics[width=1.0\linewidth]{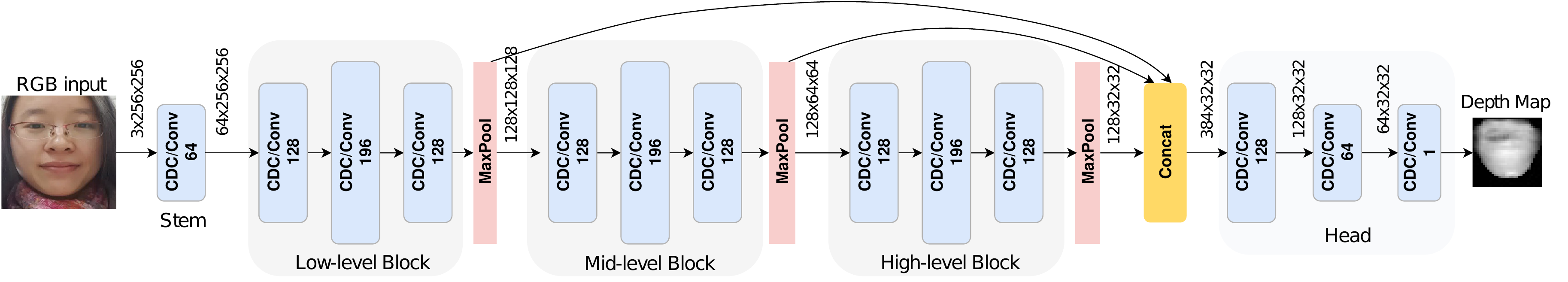}
  \caption{ 
  The multi-scale architecture of DepthNet~\cite{Liu2018Learning} with vanilla convolutions (`Conv' for short) and CDCN~\cite{yu2020searching} with CDC. Inside the blue block are the convolutional filters with 3x3 kernel size and their feature dimensionalities. 
  }
\label{fig:cdcn}
\end{figure}

\begin{figure}[t]
\centering
\includegraphics[width=0.8\linewidth]{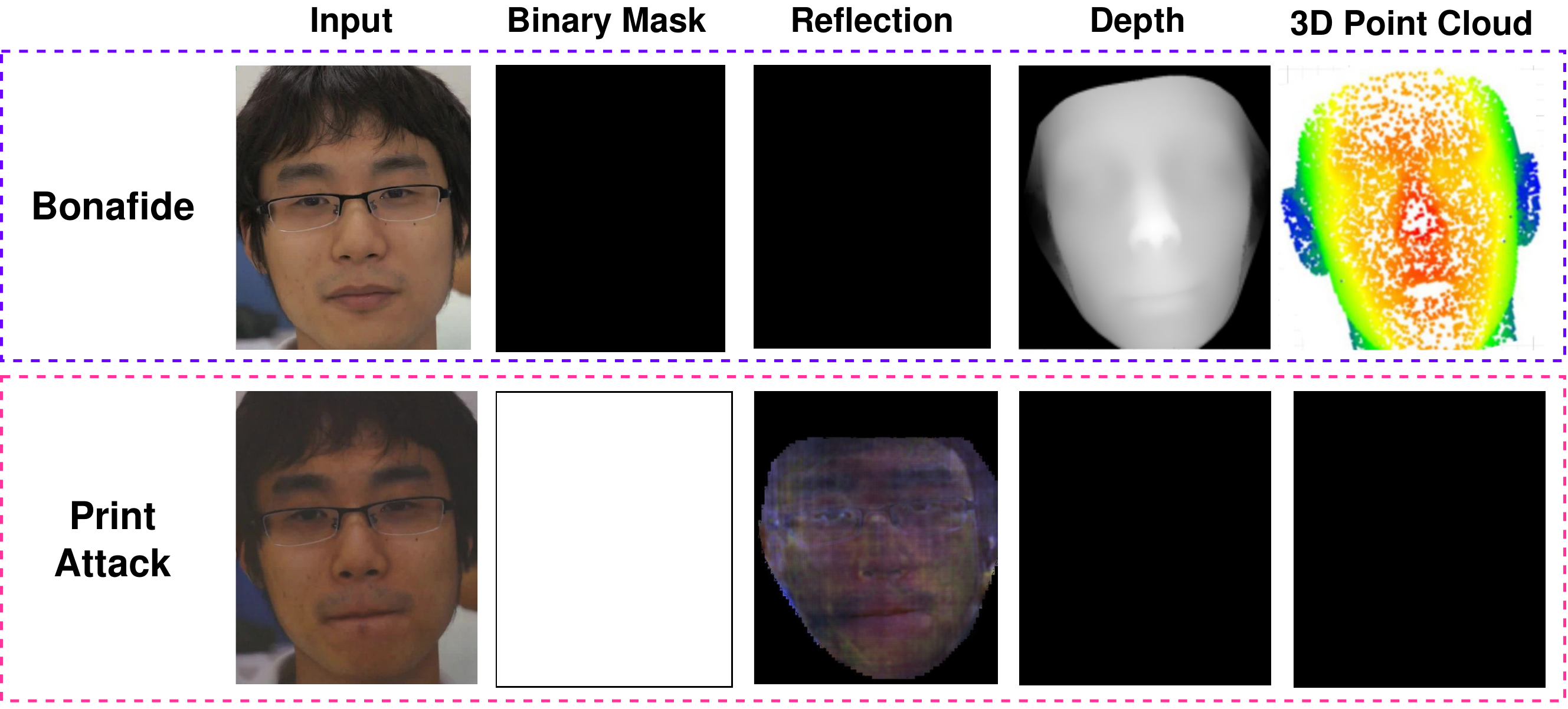}
  \caption{ 
  Visualization of pixel-wise supervision signals~\cite{yu2021revisiting} including binary mask label~\cite{george2019deep}, pseudo reflection maps~\cite{yu2020face}, pseudo depth labels~\cite{yu2020searching} and 3D point cloud maps~\cite{li3dpc} for face PAD.
  }
\label{fig:pixelwise}
\end{figure}

\vspace{0.4em}
\noindent\textbf{Traditional Deep Learning Approaches with Pixel-wise Supervision.}\quad   
Compared with bonafide faces, PAs usually have discrepant physical properties in local responses. For example, 2D PAs such as plain printed paper and electronic screen are without local geometric facial depth information while the bonafide in reverse. Motivated by this evidence, some recent works~\cite{Atoum2018Face,yu2020searching,wang2020deep} adopt pixel-wise \textit{pseudo depth} labels (see the forth column in Fig.~\ref{fig:pixelwise}) to guide the deep models, enforcing them predict the genuine depth for the bonafide samples while zero maps for the PAs. To leverage the multi-level features for accurate facial depth estimation, Atoum et al.~\cite{Atoum2018Face} propose the multi-scale fully convolutional network, namely `DepthNet'. With supervision with pseudo depth labels, the DepthNet is able to predict holistic depth maps for bonafide faces while coarse zero maps for 2D PAs as explainable decision evidence. To further improve the fine-grained intrinsic feature representation capacity, Yu et al.~\cite{yu2020searching} propose a novel deep operator called central difference convolution (CDC), which can replace vanilla convolution in DepthNet without extra learnable parameters to form the CDCN architecture (see Fig.~\ref{fig:cdcn} for detailed structures). Specifically, the CDC operator can be formulated as

\begin{equation} 
y(p_0)
=\theta \cdot \underbrace{\sum_{p_n\in R}w(p_n)\cdot (x(p_0+p_n)-x(p_0))}_{Central \ Difference \ Convolution} + (1-\theta)\cdot \underbrace{\sum_{p_n\in R}w(p_n)\cdot x(p_0+p_n)}_{Vanilla \  Convolution}, 
\label{eq:CDC}
\end{equation}
where $x$, $y$, and $w$ denote the input features, output features, and learnable convolutional weights, respectively. $p_0$ denotes the current location on both input and output feature maps while $p_n$ enumerates the locations in neighbor region $R$. The hyperparameter $\theta \in [0,1]$ trade-offs the contribution between intensity-level and gradient-level information. DepthNet with vanilla convolution is a special case of CDCN with CDC when $\theta$ = 0, i.e., aggregating local intensity information without gradient message. CDCN is favorite in pixel-wise supervision framework and widely used in the deep face PAD community due to its excellent representation capacities of both low-level detailed and high-level semantic cues. 

In consideration of the costly generation of the pseudo depth maps as well as the meaningless use for 3D face PAs with realistic depth, binary mask label~\cite{george2019deep} (see the second column in Fig.~\ref{fig:pixelwise}) is easier to be generated and more generalizable to all PAs. Specifically, binary supervision would be provided for the deep embedding features in each spatial position corresponding to the bonafide/PA distributions in each original patch (e.g., 16$\times$16). With binary mask supervision, the models are able to localize the PAs in the corresponding patches, which is attack-type-agnostic and spatially interpretable. There are also other auxiliary pixel-wise supervisions such as pseudo reflection map~\cite{yu2020face} and 3D point cloud map~\cite{li3dpc} (see the third and last columns of Fig.~\ref{fig:pixelwise}, respectively). The former provides physical material reflection cues while the latter contains denser 3D geometric cues. To further learn more intrinsic material related features, multi-head supervision are developed in~\cite{yu2020face} to supervise PAD models with multiple pixel-wise labels (i.e., pseudo depth, binary mask, and pseudo reflection)  simultaneously. The corresponding pixel-wise loss functions can be formulated as

\begin{equation} 
L_{depth}=\frac{1}{H\times W}\sum_{i\in{H},j\in{W}}\left \| D_{(i,j)} - D_{gt(i,j)}\right \|_{2}^{2},
\end{equation}
\begin{equation} 
L_{reflection}=\frac{1}{H\times W\times C}\sum_{i\in{H},j\in{W},c\in{C}}\left \| R_{(i,j,c)} - R_{gt(i,j,c)}\right \|_{2}^{2},
\end{equation}
\begin{equation} 
L_{binarymask}=\frac{1}{H\times W}\sum_{i\in{H},j\in{W}}-(B_{gt(i,j)}log(B_{(i,j)})+(1-B_{gt(i,j)})log(1-B_{(i,j)})),
\end{equation}
where $D_{gt}$, $R_{gt}$ and $B_{gt}$ denote ground truth depth map, reflection map and binary mask map respectively. $H$, $W$ and $C$ mean the height, width, and channels of the maps. Overall, pixel-wise auxiliary supervision benefits the physically meaningful and explainable representation learning. However, the pseudo auxiliary labels are usually generated coarsely without human annotations, which are sometimes inaccurate and noisy for partial attacks. For example, the binary mask for FunnyEye glasses attacks should cover the eye regions instead of the whole face).

\vspace{0.4em}
\noindent\textbf{Generalized Deep Learning Approaches to Unseen Domains.}\quad   
There might be undesired external conditional changes (e.g., in illumination and sensor quality) in real-world deployment. Traditional end-to-end deep learning based face PAD methods easily overfit to the feature distribution with training data from seen domains thus are sensitive to the domain shifts between unseen target domains and seen source domains. In the field of face PAD, `domain shifts' usually indicates the PA-irrelated external environmental changes but actually influence the appearance quality. To alleviate this issue, more recent works focus on enhancing the domain generalization capacity of the face PAD models. On the one hand, some works~\cite{shao2019multi,jia2020single} design domain-aware adversarial constraints to force the PAD models to learn domain-irrelative features from multiple source domains. They assume that the domain-irrelative features contain intrinsic bonafide/PA cues across all seen domains thus might generalize well on unseen domains. On the other hand, a few works~\cite{qin2020learning,yu2020fas2} utilize domain-aware meta-learning to learn the domain generalized feature space. Specifically, faces from partial source domains are used as query set while those from remained non-overlap domains as support set, which mimics the unseen domains and minimizes their risks at the training phase. To alternatively force the meta-learner to perform well on support sets (domains), the learned models have robust generalization capacity. Domain generalization helps the FAS model learn generalized feature representation from multiple source domains directly without any access to target data, which is more practical for real-world deployment. Despite generalization capacity enhancement for unseen domains, it would deteriorate the discrimination capability for PAD under the seen scenarios to some extent.

\subsection{Experimental Results}
\label{sec:experiment}

\begin{figure}[t]
\centering
\includegraphics[width=0.8\linewidth]{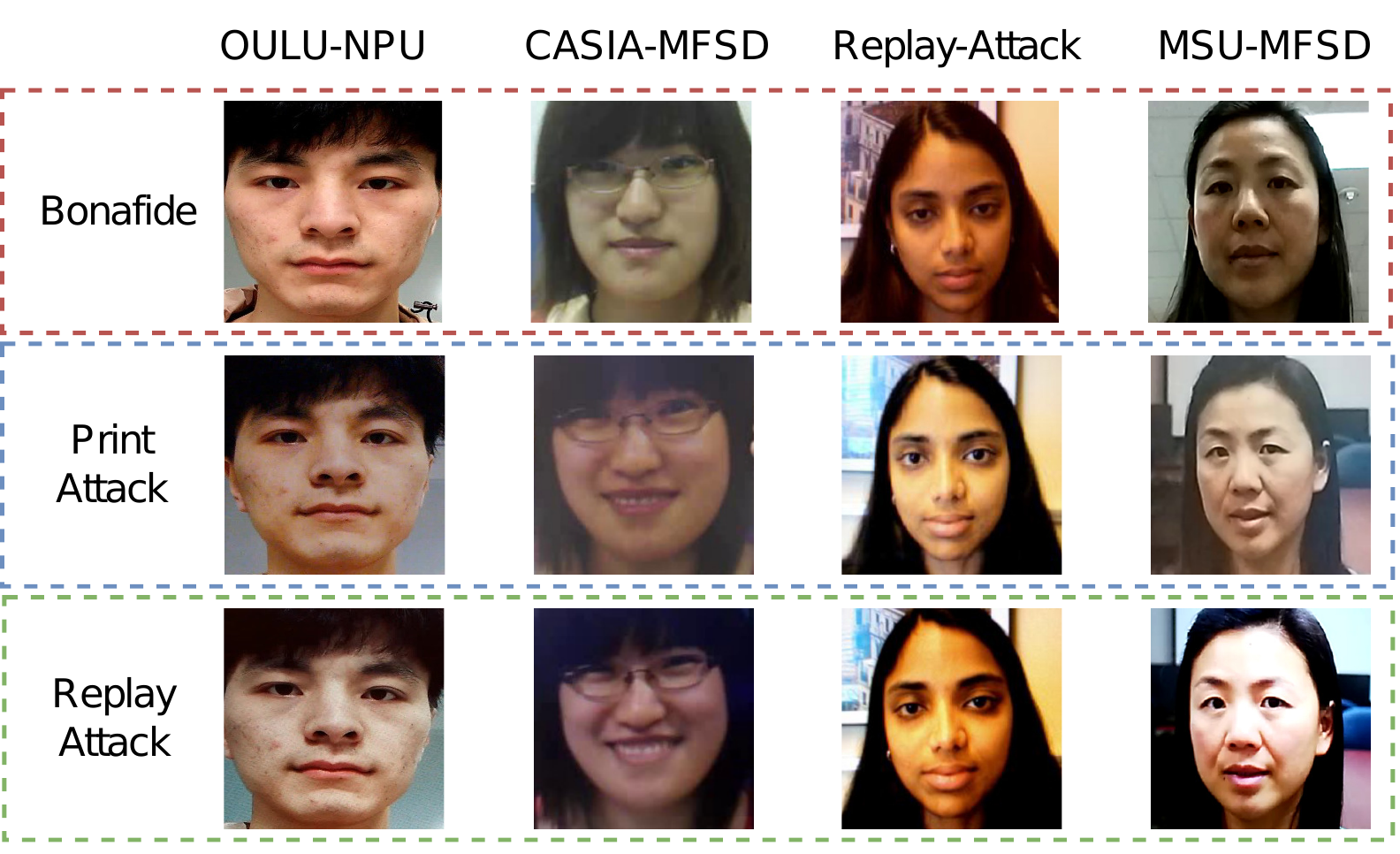}
  \caption{ 
  Visualization of the bonafide and PAs from four face PAD datasets. It can be seen that serious domain shifts (e.g., face resolution and illumination) occur among these datasets.
  }
\label{fig:crossvisual}
\end{figure}

Here, evaluation results of handcrafted feature based and deep learning based approaches on four face PAD datasets (i.e., OULU-NPU~\cite{Boulkenafet2017OULU}, CASIA-MFSD~\cite{Zhang2012A}, Replay-Attack~\cite{ReplayAttack}, and MSU-MFSD~\cite{wen2015face}) are compared and analyzed. Specifically, OULU-NPU is used for intra-dataset testings while all four datasets (see Fig.~\ref{fig:crossvisual} for typical examples) are used for cross-dataset testings under serious domain shifts.

\subsubsection{Evaluation Metrics}

As face PAD systems usually focus on the concept of bonafide and PA acceptance and rejection, two basic metrics False Rejection Rate (FRR) and False Acceptance Rate (FAR)~\cite{galbally2012high} are widely used. The ratio of incorrectly accepted spoofing attacks defines FAR, whereas FRR stands for the ratio of incorrectly rejected live accesses~\cite{chingovska2014biometrics}. The most commonly used metrics in both intra- and cross-testing scenarios is Half Total Error Rate (\textit{HTER})~\cite{chingovska2014biometrics}, Equal Error Rate (\textit{EER}), and Area Under the Curve (\textit{AUC}). HTER is found out by calculating the average of FRR (ratio of incorrectly rejected bonafide score) and FAR (ratio of incorrectly accepted PA). EER is a specific value of HTER at which FAR and FRR have equal values. AUC represents the degree of separability between bonafide and spoofings. Recently, Attack Presentation Classification Error Rate (\textit{APCER}), Bonafide Presentation Classification Error Rate (\textit{BPCER}) and Average Classification Error Rate (\textit{ACER}) suggested in ISO/IEC DIS 30107- 3:2017 standard~\cite{iso2017information} are also used for intra-dataset testings~\cite{Boulkenafet2017OULU,Liu2018Learning}. BPCER and APCER measure bonafide and attack classification error rates, respectively. ACER is calculated as the mean of BPCER and APCER, evaluating the reliability of intra-dataset performance.

\begin{table}[t]

\centering
\caption{The results of intra testing with four sub-protocols on the OULU-NPU~\cite{Boulkenafet2017OULU} dataset. The lower APCER/BPCER/ACER, the better performance. Best results are in \textbf{bold}.}

\resizebox{1.0\textwidth}{!}{
\begin{tabular}{|c|c|c|c|c|c|c|c|c|c|}
\hline
Prot.               & Method     & APCER(\%) $\downarrow$    & BPCER(\%) $\downarrow$     & ACER(\%) $\downarrow$     & Prot.               & Method     & APCER(\%) $\downarrow$     & BPCER(\%) $\downarrow$     & ACER(\%) $\downarrow$     \\ \hline
\multirow{12}{*}{1} & GRADIANT~\cite{boulkenafet2017competition}   & 1.3           & 12.5          & 6.9           & \multirow{12}{*}{2} & DeepPixBiS~\cite{george2019deep} & 11.4          & 0.6           & 6.0           \\
                    & DRL-FAS~\cite{cai2020drl}    & 5.4           & 4.0           & 4.7          &                     & De-Spoof~\cite{jourabloo2018face}   & 4.2           & 4.4           & 4.3          \\
                    & STASN~\cite{yang2019face}      & 1.2           & 2.5           & 1.9          &                     & Auxiliary~\cite{Liu2018Learning}  & 2.7           & 2.7           & 2.7          \\
                    & Auxiliary~\cite{Liu2018Learning}  & 1.6           & 1.6           & 1.6           &                     & GRADIANT~\cite{boulkenafet2017competition}   & 3.1           & 1.9           & 2.5          \\
                    & De-Spoof~\cite{jourabloo2018face}   & 1.2           & 1.7           & 1.5          &                     & Disentangled~\cite{zhang2020face}     & 1.1          & 3.6         & 2.4           \\
                    & Disentangled~\cite{zhang2020face}     & 1.7           & 0.8           & 1.3        &                      & STASN~\cite{yang2019face}      & 4.2           & 0.3           & 2.2           \\
                    &  FAS-SGTD~\cite{wang2020deep}    & 2.0           & 0.0           & 1.0         &                   & FAS-SGTD~\cite{wang2020deep}   & 2.5           & 1.3           & 1.9           \\
                    & CDCN~\cite{yu2020searching}       & 0.4           & 1.7           & 1.0         &                     & DRL-FAS~\cite{cai2020drl}    & 3.7           & \textbf{0.1}           & 1.9           \\
                    & BCN~\cite{yu2020face}        & \textbf{0.0}           & 1.6           & 0.8         &                     & BCN~\cite{yu2020face}        & 2.6           & 0.8           & 1.7           \\
                    & DeepPixBiS~\cite{george2019deep} & 0.8           & \textbf{0.0}           & 0.4          &                     & CDCN~\cite{yu2020searching}        & 1.5           & 1.4           & 1.5          \\
                    & DC-CDN~\cite{yu2021dual}     & 0.5           & 0.3           & 0.4          &                     & MT-FAS~\cite{qin2021meta}     & 1.3           & 1.4           & 1.4          \\
                    & MT-FAS~\cite{qin2021meta}     & \textbf{0.0}           & 0.8           & 0.4          &                     & DC-CDN~\cite{yu2021dual}      & \textbf{0.7}           & 1.9           & 1.3            \\
                    & NAS-FAS~\cite{yu2020fas2}    & 0.4           & \textbf{0.0}           & \textbf{0.2}       &                     & NAS-FAS~\cite{yu2020fas2}    & 1.5           & 0.8           & \textbf{1.2}         \\\hline
\multirow{12}{*}{3} & DeepPixBiS~\cite{george2019deep} & 11.7$\pm$19.6 & 10.6$\pm$14.1 & 11.1$\pm$9.4  & \multirow{12}{*}{4} & DeepPixBiS~\cite{george2019deep} & 36.7$\pm$29.7 & 13.3$\pm$14.1 & 25.0$\pm$12.7  \\
                    & GRADIANT~\cite{boulkenafet2017competition}   & 2.6$\pm$3.9   & 5.0$\pm$5.3   & 3.8$\pm$2.4  &                     & GRADIANT~\cite{boulkenafet2017competition}   & 5.0$\pm$4.5   & 15.0$\pm$7.1  & 10.0$\pm$5.0   \\
                    & De-Spoof~\cite{jourabloo2018face}   & 4.0$\pm$1.8   & 3.8$\pm$1.2   & 3.6$\pm$1.6  &                     & Auxiliary~\cite{Liu2018Learning}  & 9.3$\pm$5.6   & 10.4$\pm$6.0  & 9.5$\pm$6.0   \\
                    & DRL-FAS~\cite{cai2020drl}    & 4.6±3.6       & 1.3±1.8       & 3.0±1.5   &                     & STASN~\cite{yang2019face}      & 6.7$\pm$10.6  & 8.3$\pm$8.4   & 7.5$\pm$4.7   \\
                    & Auxiliary~\cite{Liu2018Learning}  & 2.7$\pm$1.3   & 3.1$\pm$1.7   & 2.9$\pm$1.5      &                     & DRL-FAS~\cite{cai2020drl}    & 8.1±2.7       & 6.9±5.8       & 7.2±3.9       \\
                    & STASN~\cite{yang2019face}      & 4.7$\pm$3.9   & \textbf{0.9$\pm$1.2}   & 2.8$\pm$1.6  &                     & CDCN~\cite{yu2020searching}        & 4.6±4.6       & 9.2±8.0       & 6.9±2.9 \\
                    & FAS-SGTD~\cite{wang2020deep}   & 3.2±2.0       & 2.2±1.4       & 2.7±0.6     &                     & De-Spoof~\cite{jourabloo2018face}   & 1.2$\pm$6.3   & 6.1$\pm$5.1   & 5.6$\pm$5.7       \\
                    & BCN~\cite{yu2020face}        & 2.8±2.4       & 2.3±2.8       & 2.5±1.1   &                     & BCN~\cite{yu2020face}        & 2.9±4.0       & 7.5±6.9       & 5.2±3.7   \\
                    & CDCN~\cite{yu2020searching}        & 2.4±1.3       & 2.2±2.0       & 2.3±1.4      &                     & FAS-SGTD~\cite{wang2020deep}   & 6.7±7.5       & 3.3±4.1       & 5.0±2.2      \\
                    & Disentangled~\cite{zhang2020face}     & 2.8±2.2       & 1.7±2.6       & 2.2±2.2     &                     & Disentangled~\cite{zhang2020face}     & 5.4±2.9       & 3.3±6.0       & 4.4±3.0      \\
                    & MT-FAS~\cite{qin2021meta}     & 2.3±1.5       & 1.9±1.8       & 2.1±1.7     &                     & DC-CDN~\cite{yu2021dual}      & 5.4±3.3       & 2.5±4.2       & 4.0±3.1       \\
                    & DC-CDN~\cite{yu2021dual}      & 2.2±2.8       & 1.6±2.1       & 1.9±1.1     &                     & MT-FAS~\cite{qin2021meta}     & \textbf{0.9±2.0}       & 6.4±4.9       & 3.7±2.9 \\
                    & NAS-FAS~\cite{yu2020fas2}    & \textbf{2.1$\pm$1.3}   & 1.4$\pm$1.1   & \textbf{1.7$\pm$0.6}      &                     & NAS-FAS~\cite{yu2020fas2}    & 4.2$\pm$5.3   & \textbf{1.7$\pm$2.6}   & \textbf{2.9$\pm$2.8}  \\\hline
\end{tabular}}

\label{tab:Intra}
\end{table}

\subsubsection{Intra-dataset Testings}

\begin{table}[t]

\centering
\caption{Results of the cross-dataset testings among OULU-NPU (O), CASIA-MFSD (C), Replay-Attack (I), and MSU-MFSD (M) datasets with different numbers of source domains for training. For example, `C to I' means training on CASIA-MFSD and then testing on Replay-Attack.}

\resizebox{1.0\textwidth}{!}{
\begin{tabular}{|cc|c|c|cc|cc|}
\hline
\multicolumn{2}{|c|}{\multirow{2}{*}{Method}}                                                                                & C to I   & I to C   & \multicolumn{2}{c|}{C\&O\&M to I}               & \multicolumn{2}{c|}{I\&O\&M to C}                \\ \cline{3-8} 
\multicolumn{2}{|c|}{}                                                                                                       & HTER(\%) $\downarrow$ & HTER(\%) $\downarrow$ & \multicolumn{1}{c|}{HTER(\%) $\downarrow$}      & AUC(\%) $\uparrow$        & \multicolumn{1}{c|}{HTER(\%) $\downarrow$}       & AUC(\%) $\uparrow$        \\ \hline
\multicolumn{1}{|c|}{\multirow{6}{*}{\begin{tabular}[c]{@{}c@{}}Handcrafted\\ Feature\end{tabular}}}        & LBP~\cite{Pereira2013Can}            & 55.9     & 57.6     & \multicolumn{1}{c|}{-}             & -              & \multicolumn{1}{c|}{-}              & -              \\ \cline{2-8} 
\multicolumn{1}{|c|}{}                                                                                      & Motion~\cite{Pereira2013Can}         & 50.2     & 47.9     & \multicolumn{1}{c|}{-}             & -              & \multicolumn{1}{c|}{-}              & -              \\ \cline{2-8} 
\multicolumn{1}{|c|}{}                                                                                      & LBP-TOP~\cite{Pereira2013Can}        & 49.7     & 60.6     & \multicolumn{1}{c|}{-}             & -              & \multicolumn{1}{c|}{-}              & -              \\ \cline{2-8} 
\multicolumn{1}{|c|}{}                                                                                      & Motion-Mag~\cite{tirunagari2015detection}     & 50.1     & 47.0     & \multicolumn{1}{c|}{-}             & -              & \multicolumn{1}{c|}{-}              & -              \\ \cline{2-8} 
\multicolumn{1}{|c|}{}                                                                                      & Spectral Cubes~\cite{pinto2015face} & 34.4     & 50.0     & \multicolumn{1}{c|}{-}             & -              & \multicolumn{1}{c|}{-}              & -              \\ \cline{2-8} 
\multicolumn{1}{|c|}{}                                                                                      & Color Texture~\cite{boulkenafet2015face}   & 30.3     & 37.7     & \multicolumn{1}{c|}{40.40}         & 62.78          & \multicolumn{1}{c|}{30.58}          & 76.89          \\ \hline
\multicolumn{1}{|c|}{\multirow{11}{*}{\begin{tabular}[c]{@{}c@{}}Traditional\\ Deep Learning\end{tabular}}} & De-Spoof~\cite{jourabloo2018face}       & 28.5     & 41.1     & \multicolumn{1}{c|}{-}             & -              & \multicolumn{1}{c|}{-}              & -              \\ \cline{2-8} 
\multicolumn{1}{|c|}{}                                                                                      & STASN~\cite{yang2019face}          & 31.5     & 30.9     & \multicolumn{1}{c|}{-}             & -              & \multicolumn{1}{c|}{-}              & -              \\ \cline{2-8} 
\multicolumn{1}{|c|}{}                                                                                      & Auxiliary~\cite{Liu2018Learning}      & 27.6     & 28.4     & \multicolumn{1}{c|}{29.14}         & 71.69          & \multicolumn{1}{c|}{33.52}          & 73.15          \\ \cline{2-8} 
\multicolumn{1}{|c|}{}                                                                                      & Disentangled~\cite{zhang2020face}   & 22.4     & 30.3     & \multicolumn{1}{c|}{-}             & -              & \multicolumn{1}{c|}{-}              & -              \\ \cline{2-8} 
\multicolumn{1}{|c|}{}                                                                                      & FAS-SGTD~\cite{wang2020deep}       & 17.0     & \textbf{22.8}     & \multicolumn{1}{c|}{-}             & -              & \multicolumn{1}{c|}{-}              & -              \\ \cline{2-8} 
\multicolumn{1}{|c|}{}                                                                                      & BCN~\cite{yu2020face}            & 16.6     & 36.4     & \multicolumn{1}{c|}{-}             & -              & \multicolumn{1}{c|}{-}              & -              \\ \cline{2-8} 
\multicolumn{1}{|c|}{}                                                                                      & PS~\cite{yu2021revisiting}             & 13.8     & 31.3     & \multicolumn{1}{c|}{19.55}         & 86.38          & \multicolumn{1}{c|}{18.25}          & 86.76          \\ \cline{2-8} 
\multicolumn{1}{|c|}{}                                                                                      & CDCN~\cite{yu2020searching}           & 15.5     & 32.6     & \multicolumn{1}{c|}{-}             & -              & \multicolumn{1}{c|}{-}              & -              \\ \cline{2-8} 
\multicolumn{1}{|c|}{}                                                                                      & DC-CDN~\cite{yu2021dual}         & \textbf{6.0}      & 30.1     & \multicolumn{1}{c|}{15.88}             & 91.61              & \multicolumn{1}{c|}{15.00}              & 92.80              \\ \cline{2-8} 
\multicolumn{1}{|c|}{}                                                                                      & MT-FAS~\cite{qin2021meta}          & -        & -        & \multicolumn{1}{c|}{11.93}         & 94.95          & \multicolumn{1}{c|}{18.44}          & 89.67          \\ \cline{2-8} 
\multicolumn{1}{|c|}{}                                                                                      & NAS-FAS~\cite{yu2020fas2}        & -        & -        & \multicolumn{1}{c|}{11.63}         & 96.98          & \multicolumn{1}{c|}{15.21}          & 92.64          \\ \hline
\multicolumn{1}{|c|}{\multirow{7}{*}{\begin{tabular}[c]{@{}c@{}}Generalized\\ Deep Learning\end{tabular}}}  & MADDG~\cite{shao2019multi}          & -        & -        & \multicolumn{1}{c|}{22.19}         & 84.99          & \multicolumn{1}{c|}{24.50}          & 84.51          \\ \cline{2-8} 
\multicolumn{1}{|c|}{}                                                                                      & PAD-GAN~\cite{wang2020cross}        & -        & -        & \multicolumn{1}{c|}{20.87}         & 86.72          & \multicolumn{1}{c|}{19.68}          & 87.43          \\ \cline{2-8} 
\multicolumn{1}{|c|}{}                                                                                      & RF-Meta~\cite{shao2019regularized}       & -        & -        & \multicolumn{1}{c|}{17.30}         & 90.48          & \multicolumn{1}{c|}{20.27}          & 88.16          \\ \cline{2-8} 
\multicolumn{1}{|c|}{}                                                                                      & DRDG~\cite{liu2021dual}           & -        & -        & \multicolumn{1}{c|}{15.56}         & 91.79          & \multicolumn{1}{c|}{19.05}          & 88.79          \\ \cline{2-8} 
\multicolumn{1}{|c|}{}                                                                                      & FGHV~\cite{liu2022feature}           & -        & -        & \multicolumn{1}{c|}{16.29}         & 90.11          & \multicolumn{1}{c|}{12.47}          & 93.47          \\ \cline{2-8} 
\multicolumn{1}{|c|}{}                                                                                      & SSDG~\cite{jia2020single}           & -        & -        & \multicolumn{1}{c|}{11.71}         & 96.59          & \multicolumn{1}{c|}{10.44}          & 95.94          \\ \cline{2-8} 
\multicolumn{1}{|c|}{}                                                                                      & SSAN~\cite{wang2022domain}           & -        & -        & \multicolumn{1}{c|}{\textbf{8.88}} & \textbf{96.79} & \multicolumn{1}{c|}{\textbf{10.00}} & \textbf{96.67} \\ \hline
\end{tabular}}
\label{tab:Cross}
\end{table}


Intra-dataset testing protocol has been widely used in most face PAD datasets to evaluate the model's discrimination ability for PA detection under scenarios with slight domain shift. As the training and testing data are sampled from the same datasets, they share similar domain distribution in terms of the recording environment, subject behavior, etc. The most classical intra-dataset testing protocols are the four sub-protocols of OULU-NPU dataset~\cite{Boulkenafet2017OULU}. Protocol 1 is used to evaluate the generalization performance of the face PAD algorithms under different lighting and background scenarios. Protocol 2 evaluates the PAD performance under unseen PAIs. In Protocol 3, the models are alternatively trained on videos recorded by five smartphones while videos recorded by the remaining smartphone are used for evaluation. Protocol 4 mixes the scenarios of the first three protocols to simulate real-world scenarios, and aims to evaluate the performance of face PAD methods in the integrated scenarios. The performance comparison of recent face PAD methods is shown in Table~\ref{tab:Intra}. Benefitted from the powerful representation capacity of neural networks with a data-driven fashion, most deep learning methods (except DeepPixBiS~\cite{george2019deep}) outperform the handcrafted features based method GRADIANT~\cite{george2019deep}. With the task-aware pixel-wise supervisions, some deep models such as CDCN~\cite{yu2020searching}, FAS-SGTD~\cite{wang2020deep}, Disentangled~\cite{zhang2020face}, MT-FAS~\cite{qin2021meta}, DC-CDN~\cite{yu2021dual}, and NAS-FAS~\cite{yu2020fas2} have reached satisfied performance ($\textless$5\% ACER) on the most challenging Protocol 4 with mixed domain shifts in terms of external environment, attack mediums and recording cameras.

\subsubsection{Cross-dataset Testings}

This protocol focuses on cross-dataset level domain generalization ability measurement, which usually trains models on one or several datasets (source domains) and then tests on unseen datasets (shifted target domain). We summarize recent deep face PAD approaches on two favorite cross-dataset testings~\cite{yu2020searching,shao2019multi} on four benchmark datasets (i.e., OULU-NPU (O)~\cite{Boulkenafet2017OULU}, CASIA-MFSD (C)~\cite{Zhang2012A}, Replay-Attack (I)~\cite{ReplayAttack}, and MSU-MFSD (M)~\cite{wen2015face}) in Table~\ref{tab:Cross}. As illustrated in Fig.~\ref{fig:crossvisual}, there are serious domain shifts among these four datasets in terms of resolution, illumination, sensor noise, etc. 
When trained on Replay-Attack and tested on CASIA-MFSD, most handcrafted feature based methods as well as traditional deep models perform poorly ($\textgreater$20\% HTER) due to the serious lighting and camera resolution variations. In contrast, when trained on multiple source datasets (i.e., OULU-NPU, MSU-MFSD, and Replay-Attack), domain generalization based methods achieve acceptable performance on CASIA-MFSD (especially SSDG~\cite{jia2020single} and SSAN~\cite{wang2022domain} with 10.44\% and 10.00\% HTER, respectively). Overall, introducing more training data from diverse domains might benefit and stabilize the generalized feature learning.


\subsection{Applications}
\label{sec:Applications}
In this subsection, we will concentrate on describing face presentation attack detection in different industry applications, including the attributes of the scenario, sensors, protocols and approaches.




\subsubsection{Online Identity Verification Scenario}
\begin{figure} [t]
\centering
\includegraphics[width=1.0\linewidth]{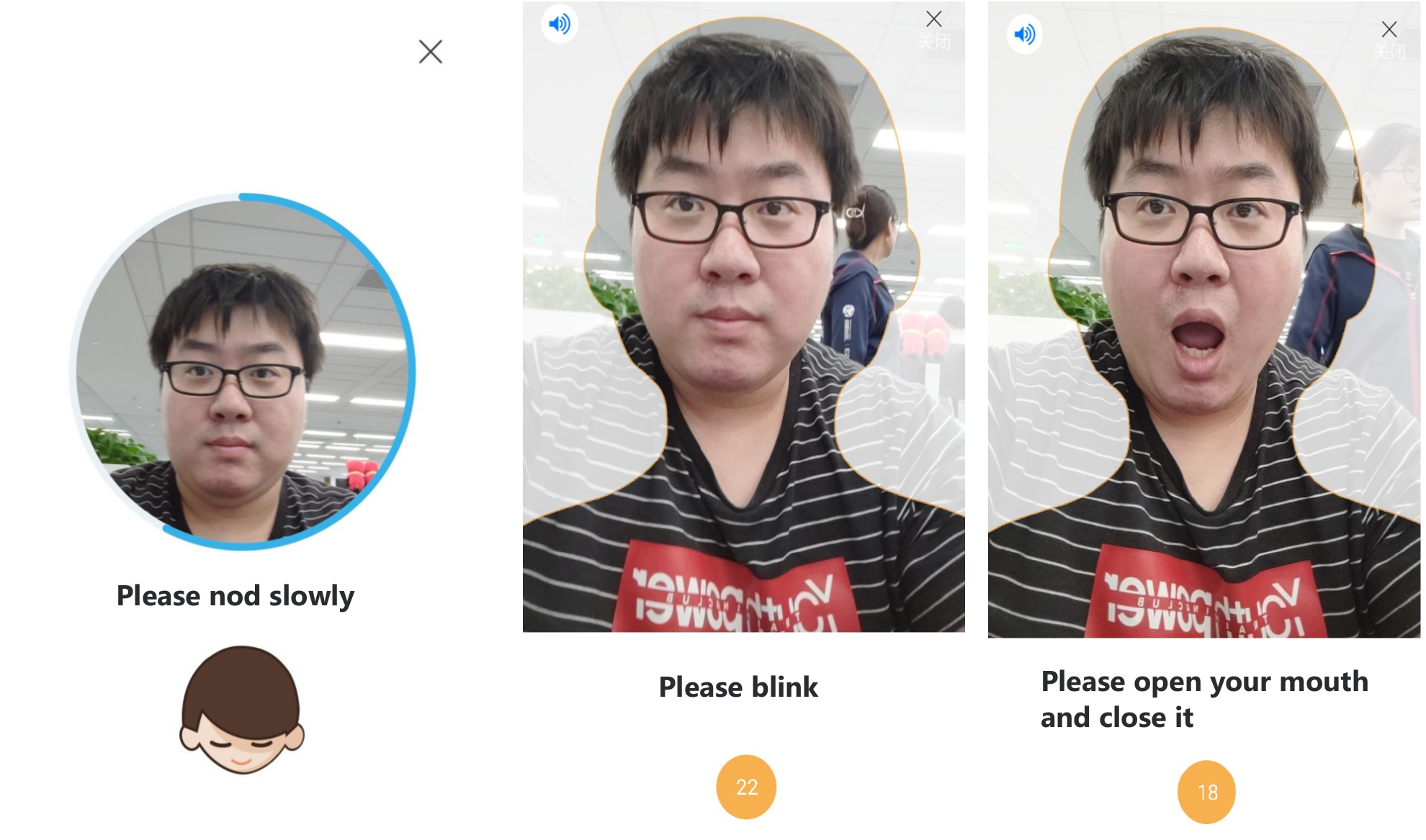}
  \caption{ 
  The online identity verification scenario. The customer completes an identity authentication process online through the mobile APP. During this process, it is usually required to make cooperative actions according to the system prompts.
  }
\label{fig:scenario1}
\end{figure}

As illustrated in Fig.\ref{fig:scenario1}, this scenario refers to the online face recognition authentication process by customers through their mobile devices or PCs. Face PAD in the online identity verification scenario aims to force the algorithm discriminate spoofing faces from the criminals. Criminals attempt to obtain the authentication result of the attacked individual via spoofing faces. After obtaining the authentication result, they can steal the money or information from the accessed account. The architectures of this scenario are:
\begin{itemize}
\item The system requires a high level of security indicating the face PAD algorithm requires to reach a higher performance.
\item The application runs on the client side, and most of the devices are mobile phones. Thus, the algorithm needs to reach strong hardware compatibility.
\item The criminals' attack conditions are more relaxed due to the relatively private application environment. Alternatively, the attack cost is cheap, and repeated attempts can be made.
\item A variety of devices and unpredictable novel
PAs keep evolving and unknown PAs may be presented to them. Data-driven models may give unpredictable results when faced with out-of-distribution samples.
\item Customers could cooperate to a certain extent.
\item Only one face is allowed in one operation process, and multiple faces will be regarded as illegal operations.

\end{itemize}

\textbf{Sensors} in this scenario are usually diverse due to the customers' diverse mobile devices. In order to achieve satisfactory compatibility for the diverse hardware, we prefer the RGB modality because most hardware devices support this modality. However, approaches designed with the single RGB modality usually have weak feature representation compared with multimodal inputs. To bridge this gaps, large-scale training data would be collected to cover as many PAs and domains as possible.

\begin{figure} [t]
\centering
\includegraphics[width=1.0\linewidth]{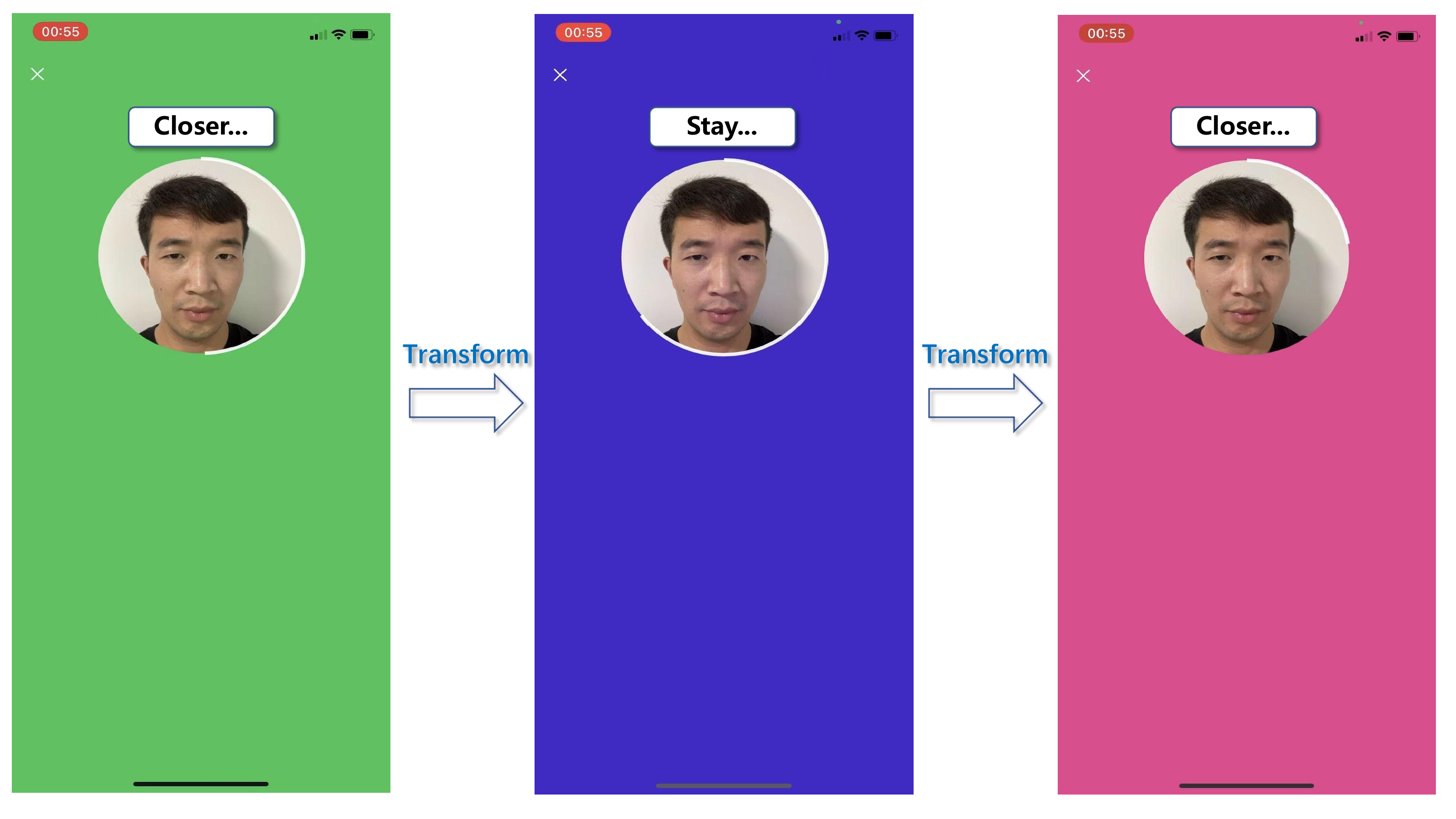}
  \caption{ 
  Color verification code. During one verification process, verifying whether the reflective color matches the color verification code in which the background presents high-brightness images of different colors provides additional dimensional knowledge for the face PAD algorithm.
  }
\label{fig:guard}
\end{figure}

\textbf{Approaches} in this scenario always treat the face PAD problem as a binary classification task~\cite{Li2017An}, and utilize binary cross-entropy loss to optimize the model. Domain adaptation and generalization
approaches~\cite{li2018domain, jia2020single,  wang2020cross} can also be applied in this scenario. For example, meta-learning~\cite{wang2021self,liu2021dual,liu2021adaptive} based methods can be adopted to improve the model’s generalization capacity on unseen attacks and domains. To enhance the robustness, in this scenario, the face PAD system usually receives additional dynamic information by requiring the customers to cooperate to complete the facial actions~\cite{li2008eye,wang2009face} or by changing the color of the screen~\cite{zhang2021aurora} (see Fig.\ref{fig:guard} for visualization). This interactive instructions is also called \textit{Liveness Detection.}

\vspace{0.4em}
\noindent\textbf{Protocols and Evaluations}. For better evaluation under this scenario, we need to build a sufficiently large dataset and a sufficiently rigorous testing protocol. Considering that the attacker in this scenario can repeatedly attack the system in a private environment, we need to ensure that the PAD algorithm can reach a high true positive rate for PAs and decrease the false negative rate based on all bonafide samples being correctly detected. To further verify this goal, the Receiver Operating Characteristic (ROC) curve~\cite{bi2003regression} is proposed to evaluate the face PAD method on the large-scale dataset. In the ROC, we pursue a lower false positive rate and higher true negative rate under the same false positive rate.

\vspace{0.4em}
\noindent\textbf{Related Applications}. In this scenario, there are some typical applications with similar characteristics, such as the FaceID of mobile devices. In contrast, mobile phone manufacturers can select more sensors, some of which have multiple modalities, which could improve the security level.

\subsubsection{Offline Payment Scenario}
\begin{figure} [t]
\centering
\includegraphics[width=1.0\linewidth]{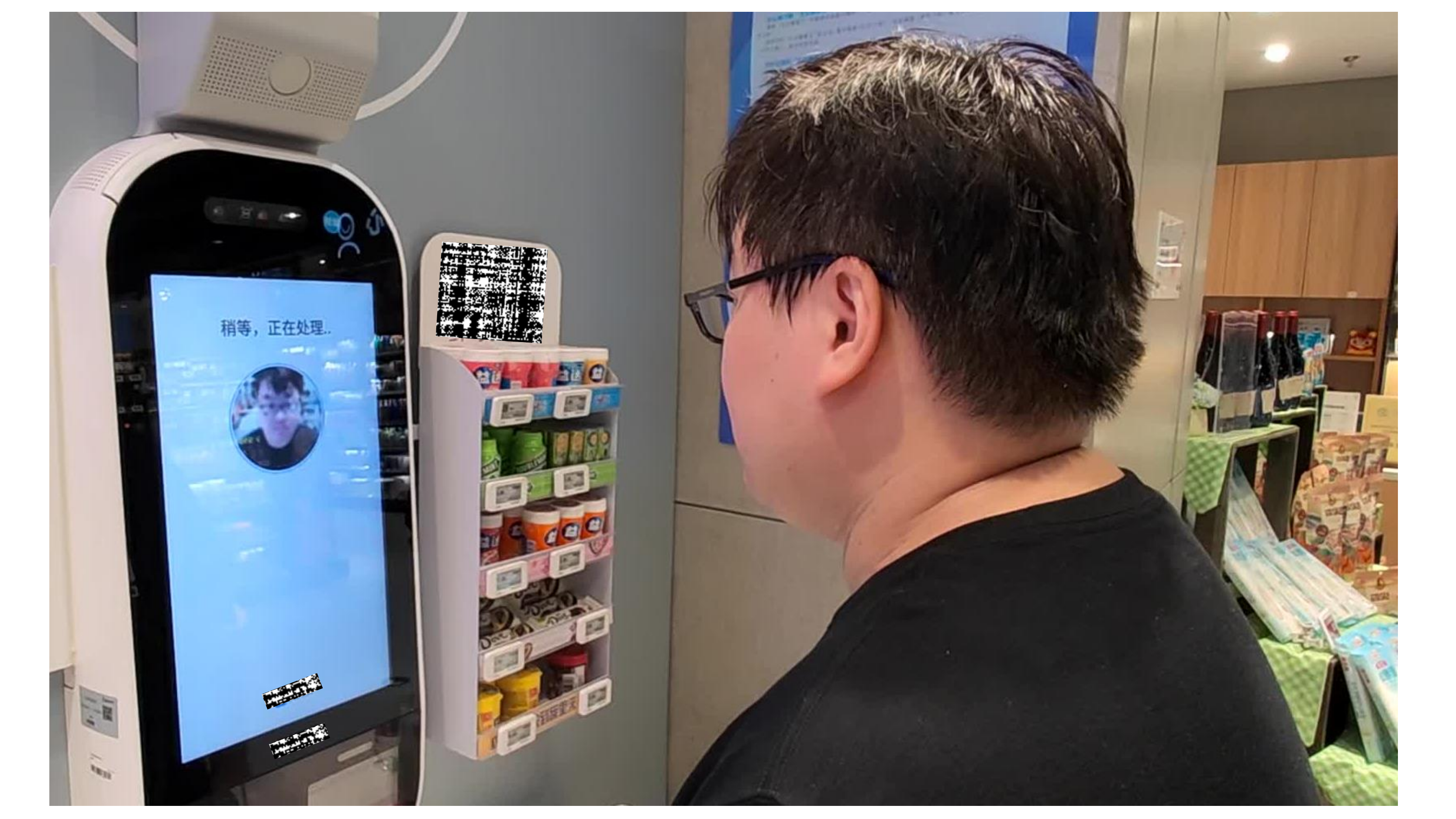}
  \caption{ 
  The offline payment scenario. The customer completes a face recognition payment process through an industry-specific machine with a fixed offline location. Such special machines are generally equipped with multi-modalities sensors.
  }
\label{fig:scenario2}
\end{figure}

As illustrated in Fig.~\ref{fig:scenario2}, this scenario refers to the process in which the customer utilizes the fixed face recognition instrument for offline identity authentication or payment. Face PAD approach aims to secure the face recognition system from malicious PAs including 3D masks. This scenario has the following diverse characteristics:
\begin{itemize}
\item Offline payment scenarios will directly involve money transactions, which require the system to dedicate a very high-security level, which also requires the face anti-spoofing algorithm performance to reach a higher level. In terms of this, a single RGB modality is almost incapable.
\item The application also runs on the client side. However, the carrier equipment of the system is generally a standardized industry-specific machine, and multimodal cameras can be equipped. The domain in this scenario is relatively simple due to the device's fixed sensor and fixed location.
\item The criminals' attack conditions are constrained due to the relatively public application environment and fixed device location. Generally, there will be staff or other customers on site. For attackers, the attack cost increases as the numbers of repeated attempts are reduced.
\item In this scenario, the most significant challenge to the system comes from the 3D high-fidelity masks and head models. Because equipped with multimodal cameras, which effectively defends the common planar PAs such as print, replay and paper mask.
\item Customers are only required to do limited cooperation.
\item Only one face is allowed in one operation process, and multiple faces will be regarded as illegal operations.

\end{itemize}

\begin{figure} [t]
\centering
\includegraphics[width=1.0\linewidth]{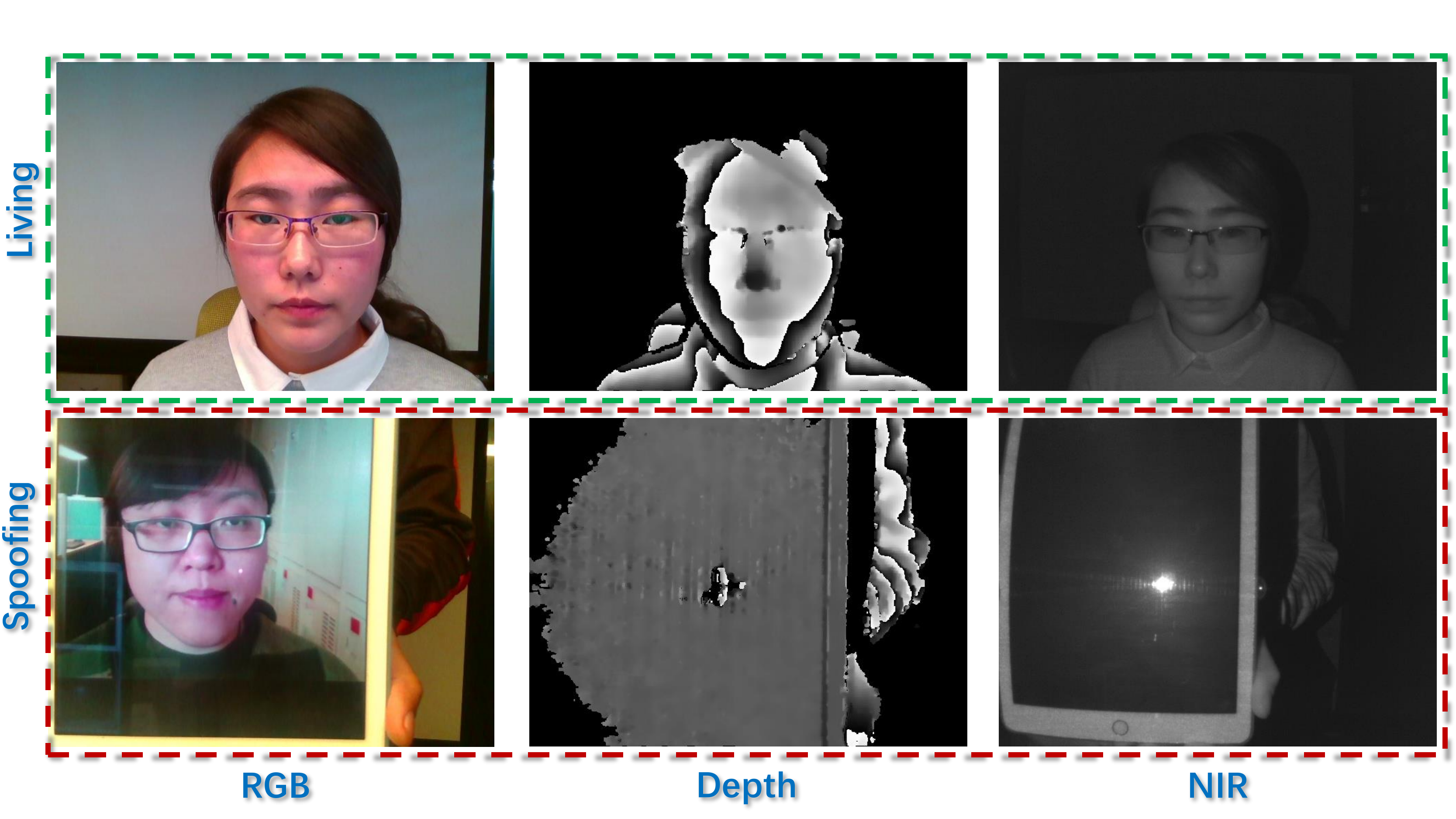}
  \caption{ 
  Imaging of bonafide and spoofing faces in RGB, NIR and Depth multimodal sensors. In contrast, the 2D planar PAs show significantly different patterns in imaging NIR and Depth.
  }
\label{fig:Multi-modalities}
\end{figure}

\textbf{Sensors} in this scenario prefer to choose multimodal cameras, such as RGB and NIR binocular camera or RGB, NIR and Depth structured light camera~\cite{geng2011structured} or TOF camera~\cite{foix2011lock}. The combination of NIR and Depth modalities aims to effectively defend against planar attacks such as print and replay. As illustrated in Fig.~\ref{fig:Multi-modalities}, the 2D planar PAs cannot perform face imaging in these two modalities. Combined with the RGB modality, it can defend against some 3D forms of attack, such as 3D high-fidelity masks.

\textbf{Approaches} in this scenario treated the face PAD problem as a typical multimodal fusion task. With multi-modal inputs, mainstream methods extract complementary multimodal features using feature-level fusion strategies. As there are redundancy across multimodal features, direct feature concatenation easily results in high-dimensional features and overfitting. To alleviate this issue, Zhang et al.~\cite{casiasurf} propose the SD-Net utilize a feature re-weighting mechanism to select the informative and discard the redundant channel features among RGB, depth, and NIR modalities. However, even if the features of the three modalities are combined, some spoofing faces are still challenging to discriminate, such as 3D high-fidelity masks. To further boost the multi-modal performance, Liu et al.~\cite{9813717} propose a large-scale 3D high-fidelity mask dataset and the contrastive context-aware learning, which is able to learn by leveraging rich contexts accurately among pairs of bonafide and high-fidelity mask attack.

\vspace{0.4em}
\noindent\textbf{Protocols and Evaluations}. A sufficiently large-scale dataset as well as rigorous testing protocols should be established to evaluate the algorithm for this scenario. Considering that the system in this scenario require a very high-security level, we need to ensure that the algorithm can reach a high true positive rate for the spoofing faces and decrease the false negative rate based on all positive samples being correctly detected. ROC curve~\cite{bi2003regression} is utilized to evaluate the face PAD method on the large-scale dataset. In the ROC, we pursue a lower false positive rate and higher true negative rate under the same false positive rate.

\vspace{0.4em}
\noindent\textbf{Related Applications}. In this scenario, there are some typical applications with similar characteristics, such as face access control in the buildings. Face access control has relatively low requirements on the system's security level, and different sensors can be selected according to the actual condition and cost.

\begin{figure} [t]
\centering
\includegraphics[width=1.0\linewidth]{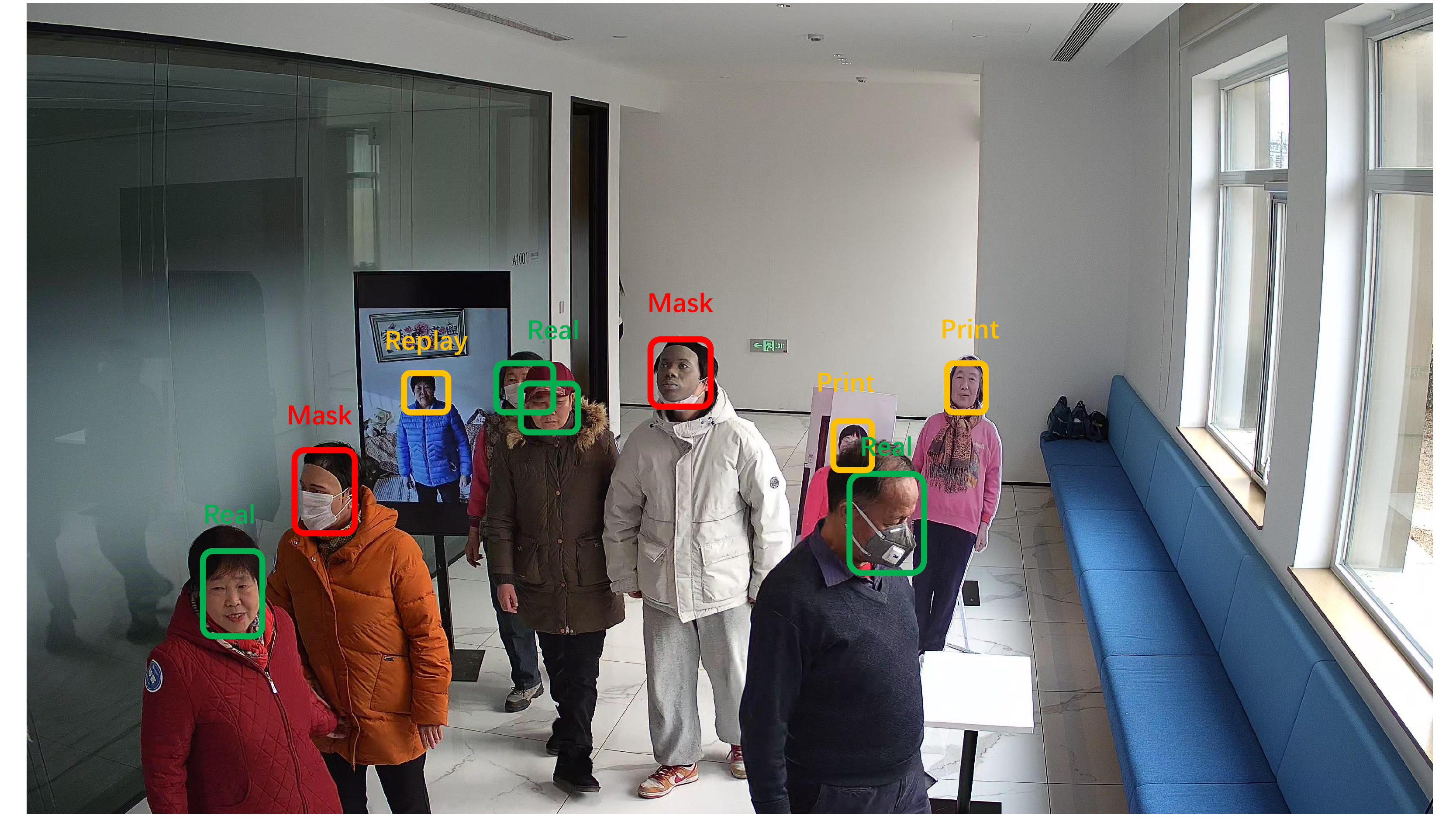}
  \caption{ 
  The surveillance scenario. There will be multiple faces involved in the surveillance camera. In addition to regular faces, there will be criminals wearing high-fidelity masks and faces that belong to noise in the background of posters, advertisements, etc.
  }
\label{fig:scenario3}
\end{figure}

\subsubsection{Surveillance Scenario}
As illustrated in Fig.\ref{fig:scenario3}, this scenario refers to the process of the customer unconsciously passing through the surveillance camera framing area. Compared with the above two scenarios, the function of the face anti-spoofing module in the surveillance scenario is quite different. The PAs in this scenario are mainly divided into two categories. One is the criminal who wears a mask and mixes in the crowd trying to escape the surveillance. The other is the PAs shown on the screen or demonstrated on the billboard in the background of the surveillance. For the monitoring system, the first category is an attack behavior, and the second category is a noise. This scenario has the following diverse characteristics:
\begin{itemize}
\item According to the properties of the above-mentioned two categories of face PAs, the face PAD approach in this scenario aims to caution against abnormal behavior and remove background noise, so it means nothing to the system requires a very high-security level.
\item The application can run on the cloud side. Because this is surveillance or a similar scenario, the sensors deployed are mostly surveillance cameras.
\item The imaging in this scenario is a kind of long-range monitoring. In the long-range monitoring, the features of each low-resolution face are relatively sparse, which increases the difficulty for the face PAD algorithm. Alternatively, in order to be caught by the surveillance cameras as less as possible, the criminals will pass through the acquisition area of the cameras as quickly as possible. In other words, the number of repeated attacks is reduced.
\item In long-range camera monitoring, the most challenging problem in this scenario is whether the face PAD algorithm could effectively discriminate bonafide faces, 3D high-fidelity masks, and print/replay in the background.
\item The customers are completely passive and do not require to cooperate.
\item The number of faces is no longer a limitation of the system.

\end{itemize}

\textbf{Sensors} deployed in this scenario are mostly surveillance cameras~\cite{qureshi2006surveillance}. The images captured by the cameras with long imaging distance and wide imaging range. Multiple faces emerge in the viewfinder at the same time and at different distances. It also includes faces on screens and posters. Some of these sensors also contain multimodal modules, and these multimodal cameras (such as NIR) can mainly deal with dark light environments. In fact, due to the long distance, it is challenging to capture rich spoofing cues. Alternatively, in this scenario, only the RGB modality can provide adequate embedding knowledge.

\textbf{Approaches} in this scenario formulate a long-distance face PAD problem. To the best of our knowledge, this issue is less studied in the current face PAD community, and we still do not have a standardized, well-defined benchmark.

\vspace{0.4em}
\noindent\textbf{Protocols and Evaluations}. To better evaluate the algorithm for this scenario, we need to establish a sufficiently large-scale dataset as well as generalized protocols. Considering that the system in this scenario does not require a very high-security level, we firstly ensure that the algorithm can reach a high true negative rate and bonafide faces will not be misidentified, making the system repeatedly alarm. Based on this foundation, we will continue to decrease the false negative rate. To further verify this goal, the ROC curve~\cite{bi2003regression} is utilized to evaluate the face PAD method on the large-scale dataset. In the ROC, we pursue a lower false negative rate and higher true positive rate under the same false negative rate.


\vspace{0.4em}
\noindent\textbf{Related Applications}. In this scenario, there are some typical applications with similar characteristics, such as the passenger flow monitoring in the market. In contrast, face PAD approaches in this type of system only require excluding the face on the poster or screen.

\subsection{Conclusion and Future Challenge}
\label{sec:conclusion}


In this chapter, a comprehensive review of the research and practical applications related to face PAD are carried out. On the one hand, handcrafted feature and deep learning based approaches based on unimodal or multimodal cameras can all detect face PAs to a certain extent without rigorous interaction. It is still hard to say which feature performs better in different applicational scenarios. For example, traditional rPPG based methods are good at 3D mask attack detection in unimodal RGB scenarios while appearance-based deep learning methods perform well on 2D face PAD. Extensive experiments and practical applications have proved that 1) combining multiple features can achieve better face PAD results; and 2) task-aware prior knowledge (e.g., pixel-wise supervision), advanced learning paradigms (e.g., domain generalization) and multimodal inputs (e.g., RGB-NIR-Depth) can benefit the discrimination, generalization and interpretability of face PAD algorithms. On the other hand, hardware cost of multimodal sensors (e.g., RGB+Depth+NIR is more expensive and costs more spatial spaces than RGB+Depth or RGB+NIR) and practical application requirements (e.g., distance and real-time efficiency) need to be comprehensively considered for real-world deployment under different applicational scenarios. 


Face PAD has achieved rapid improvement over the past few years due to advanced camera sensors and well-designed algorithms. However, face PAD is still an unsolved problem in terms of task-oriented challenges such as subtle spoof pattern representation, complex real-world domain gaps, and rapidly iterative novel attacks as well as novel applicational scenarios like long-range monitoring for surveillance. Here we list the two limitations of the current development. On one side, the evaluation under saturating and unpractical testing benchmarks/protocols cannot really reflect the effectiveness of the methods for real-world applications. For example, most datasets are recorded in controlled lab environments but rarely considering the real-world offline payment and surveillance scenarios. Thus, it is urgent to establish more large-scale practical application-oriented benchmarks to bridge the gaps between academia and industry. On the other side, most existing works train and adapt deep face PAD models with huge stored source face data in fixed scenarios, and neglect 1) the privacy and biometric sensitivity issue; and 2) the continuous adaptation for emerging domains and attacks. To design source-free continuous learning face PAD algorithms might be potential for domain-robust real-world deployment and dynamic updating against novel attacks.


\begin{acknowledgement}
This work was supported by the National Natural Science Foundation of China (No. 62276254, 62176256, 62206280 and 62106264) and the InnoHK program.
\end{acknowledgement}

\ifthenelse{\equal{false}{\buildbook}}{
\printindex
\printglossary
\bibliographystyle{spmpsci}
\bibliography{references}

\begin{thebibliography}{10}
\providecommand{\url}[1]{{#1}}
\providecommand{\urlprefix}{URL }
\expandafter\ifx\csname urlstyle\endcsname\relax
  \providecommand{\doi}[1]{DOI~\discretionary{}{}{}#1}\else
  \providecommand{\doi}{DOI~\discretionary{}{}{}\begingroup
  \urlstyle{rm}\Url}\fi

\bibitem{agarwal2017face}
Agarwal, A., Yadav, D., Kohli, N., Singh, R., Vatsa, M., Noore, A.: Face
  presentation attack with latex masks in multispectral videos.
\newblock In: CVPRW (2017)

\bibitem{ali2012liveness}
Ali, A., Deravi, F., Hoque, S.: Liveness detection using gaze collinearity.
\newblock In: ICEST. IEEE (2012)

\bibitem{Atoum2018Face}
Atoum, Y., Liu, Y., Jourabloo, A., Liu, X.: Face anti-spoofing using patch and
  depth-based cnns.
\newblock In: IJCB (2017)

\bibitem{bhattacharjee2018spoofing}
Bhattacharjee, S., Mohammadi, A., Marcel, S.: Spoofing deep face recognition
  with custom silicone masks.
\newblock In: BTAS (2018)

\bibitem{bi2003regression}
Bi, J., Bennett, K.P.: Regression error characteristic curves.
\newblock In: ICML (2003)

\bibitem{iso2017information}
Biometrics., I.J.S.: Information technology--biometric presentation attack
  detection--part 3: testing and reporting  (2017)

\bibitem{boulkenafet2017competition}
Boulkenafet, Z., Komulainen, J., Akhtar, Z., Benlamoudi, A., Samai, D.,
  Bekhouche, S.E., Ouafi, A., Dornaika, F., Taleb-Ahmed, A., Qin, L., et~al.: A
  competition on generalized software-based face presentation attack detection
  in mobile scenarios.
\newblock In: IJCB. IEEE (2017)

\bibitem{boulkenafet2015face}
Boulkenafet, Z., Komulainen, J., Hadid, A.: Face anti-spoofing based on color
  texture analysis.
\newblock In: ICIP (2015)

\bibitem{Boulkenafet2016Face}
Boulkenafet, Z., Komulainen, J., Hadid, A.: Face spoofing detection using
  colour texture analysis.
\newblock TIFS  (2016)

\bibitem{Boulkenafet2017OULU}
Boulkenafet, Z., Komulainen, J., Li, L., Feng, X., Hadid, A.: Oulu-npu: A
  mobile face presentation attack database with real-world variations.
\newblock In: FGR (2017)

\bibitem{cai2020drl}
Cai, R., Li, H., Wang, S., Chen, C., Kot, A.C.: Drl-fas: A novel framework
  based on deep reinforcement learning for face anti-spoofing.
\newblock IEEE TIFS  (2020)

\bibitem{chen2019attention}
Chen, H., Hu, G., Lei, Z., Chen, Y., Robertson, N.M., Li, S.Z.: Attention-based
  two-stream convolutional networks for face spoofing detection.
\newblock TIFS  (2019)

\bibitem{ReplayAttack}
Chingovska, I., Anjos, A., Marcel, S.: On the effectiveness of local binary
  patterns in face anti-spoofing.
\newblock In: Biometrics Special Interest Group (2012)

\bibitem{chingovska2014biometrics}
Chingovska, I., Dos~Anjos, A.R., Marcel, S.: Biometrics evaluation under
  spoofing attacks.
\newblock IEEE TIFS  (2014)

\bibitem{costa2016replay}
Costa-Pazo, A., Bhattacharjee, S., Vazquez-Fernandez, E., Marcel, S.: The
  replay-mobile face presentation-attack database.
\newblock In: BIOSIG. IEEE (2016)

\bibitem{de2012moving}
De~Marsico, M., Nappi, M., Riccio, D., Dugelay, J.L.: Moving face spoofing
  detection via 3d projective invariants.
\newblock In: ICB, pp. 73--78. IEEE (2012)

\bibitem{erdogmus2014spoofing}
Erdogmus, N., Marcel, S.: Spoofing face recognition with 3d masks.
\newblock TIFS  (2014)

\bibitem{foix2011lock}
Foix, S., Alenya, G., Torras, C.: Lock-in time-of-flight (tof) cameras: A
  survey.
\newblock IEEE Sensors Journal \textbf{11}(9), 1917--1926 (2011)

\bibitem{Pereira2013Can}
de~Freitas~Pereira, T., Anjos, A., De~Martino, J.M., Marcel, S.: Can face
  anti-spoofing countermeasures work in a real world scenario?
\newblock In: ICB (2013)

\bibitem{galbally2012high}
Galbally, J., Alonso-Fernandez, F., Fierrez, J., Ortega-Garcia, J.: A high
  performance fingerprint liveness detection method based on quality related
  features.
\newblock Future Generation Computer Systems \textbf{28}(1), 311--321 (2012)

\bibitem{galbally2013image}
Galbally, J., Marcel, S., Fierrez, J.: Image quality assessment for fake
  biometric detection: Application to iris, fingerprint, and face recognition.
\newblock IEEE TIP  (2013)

\bibitem{geng2011structured}
Geng, J.: Structured-light 3d surface imaging: a tutorial.
\newblock Advances in Optics and Photonics \textbf{3}(2), 128--160 (2011)

\bibitem{george2019deep}
George, A., Marcel, S.: Deep pixel-wise binary supervision for face
  presentation attack detection.
\newblock In: ICB (2019)

\bibitem{george2020effectiveness}
George, A., Marcel, S.: On the effectiveness of vision transformers for
  zero-shot face anti-spoofing.
\newblock arXiv preprint arXiv:2011.08019  (2020)

\bibitem{george2019biometric}
George, A., Mostaani, Z., Geissenbuhler, D., Nikisins, O., Anjos, A., Marcel,
  S.: Biometric face presentation attack detection with multi-channel
  convolutional neural network.
\newblock TIFS  (2019)

\bibitem{heusch2020deep}
Heusch, G., George, A., Geissb{\"u}hler, D., Mostaani, Z., Marcel, S.: Deep
  models and shortwave infrared information to detect face presentation
  attacks.
\newblock TBIOM  (2020)

\bibitem{jia20203d}
Jia, S., Li, X., Hu, C., Guo, G., Xu, Z.: 3d face anti-spoofing with factorized
  bilinear coding.
\newblock arXiv preprint arXiv:2005.06514  (2020)

\bibitem{jia2020single}
Jia, Y., Zhang, J., Shan, S., Chen, X.: Single-side domain generalization for
  face anti-spoofing.
\newblock In: CVPR (2020)

\bibitem{jourabloo2018face}
Jourabloo, A., Liu, Y., Liu, X.: Face de-spoofing: Anti-spoofing via noise
  modeling.
\newblock In: ECCV (2018)

\bibitem{Komulainen2014Context}
Komulainen, J., Hadid, A., Pietikainen, M.: Context based face anti-spoofing.
\newblock In: BTAS (2013)

\bibitem{li2020casia}
Li, A., Tan, Z., Li, X., Wan, J., Escalera, S., Guo, G., Li, S.Z.: Casia-surf
  cefa: A benchmark for multi-modal cross-ethnicity face anti-spoofing.
\newblock WACV  (2021)

\bibitem{li2018domain}
Li, H., Jialin~Pan, S., Wang, S., Kot, A.C.: Domain generalization with
  adversarial feature learning.
\newblock In: CVPR (2018)

\bibitem{li2018unsupervised}
Li, H., Li, W., Cao, H., Wang, S., Huang, F., Kot, A.C.: Unsupervised domain
  adaptation for face anti-spoofing.
\newblock IEEE TIFS  (2018)

\bibitem{li2008eye}
Li, J.W.: Eye blink detection based on multiple gabor response waves.
\newblock In: ICMLC, vol.~5, pp. 2852--2856. IEEE (2008)

\bibitem{Li2017An}
Li, L., Feng, X., Boulkenafet, Z., Xia, Z., Li, M., Hadid, A.: An original face
  anti-spoofing approach using partial convolutional neural network.
\newblock In: IPTA (2016)

\bibitem{li2020compactnet}
Li, L., Xia, Z., Jiang, X., Roli, F., Feng, X.: Compactnet: learning a compact
  space for face presentation attack detection.
\newblock Neurocomputing  (2020)

\bibitem{li2016generalized}
Li, X., Komulainen, J., Zhao, G., Yuen, P.C., Pietik{\"a}inen, M.: Generalized
  face anti-spoofing by detecting pulse from face videos.
\newblock In: ICPR (2016)

\bibitem{li3dpc}
Li, X., Wan, J., Jin, Y., Liu, A., Guo, G., Li, S.Z.: 3dpc-net: 3d point cloud
  network for face anti-spoofing  (2020)

\bibitem{9813717}
Liu, A., Zhao, C., Yu, Z., Wan, J., Su, A., Liu, X., Tan, Z., Escalera, S.,
  Xing, J., Liang, Y., Guo, G., Lei, Z., Li, S.Z., Zhang, D.: Contrastive
  context-aware learning for 3d high-fidelity mask face presentation attack
  detection.
\newblock IEEE TIFS  (2022)

\bibitem{liu2021contrastive}
Liu, A., Zhao, C., Yu, Z., Wan, J., Su, A., Liu, X., Tan, Z., Escalera, S.,
  Xing, J., Liang, Y., et~al.: Contrastive context-aware learning for 3d
  high-fidelity mask face presentation attack detection.
\newblock IEEE TIFS  (2022)

\bibitem{liu2020temporal}
Liu, S., Lan, X., Yuen, P.: Temporal similarity analysis of remote
  photoplethysmography for fast 3d mask face presentation attack detection.
\newblock In: WACV (2020)

\bibitem{liu2022feature}
Liu, S., Lu, S., Xu, H., Yang, J., Ding, S., Ma, L.: Feature generation and
  hypothesis verification for reliable face anti-spoofing.
\newblock In: AAAI (2022)

\bibitem{liu20163d}
Liu, S., Yuen, P.C., Zhang, S., Zhao, G.: 3d mask face anti-spoofing with
  remote photoplethysmography.
\newblock In: ECCV. Springer (2016)

\bibitem{liu2021adaptive}
Liu, S., Zhang, K.Y., Yao, T., Bi, M., Ding, S., Li, J., Huang, F., Ma, L.:
  Adaptive normalized representation learning for generalizable face
  anti-spoofing.
\newblock In: ACM MM (2021)

\bibitem{liu2021dual}
Liu, S., Zhang, K.Y., Yao, T., Sheng, K., Ding, S., Tai, Y., Li, J., Xie, Y.,
  Ma, L.: Dual reweighting domain generalization for face presentation attack
  detection.
\newblock In: IJCAI (2021)

\bibitem{Liu2018Learning}
Liu, Y., Jourabloo, A., Liu, X.: Learning deep models for face anti-spoofing:
  Binary or auxiliary supervision.
\newblock In: CVPR (2018)

\bibitem{liu2019deep}
Liu, Y., Stehouwer, J., Jourabloo, A., Liu, X.: Deep tree learning for
  zero-shot face anti-spoofing.
\newblock In: CVPR (2019)

\bibitem{maatta2011face}
M{\"a}{\"a}tt{\"a}, J., Hadid, A., Pietik{\"a}inen, M.: Face spoofing detection
  from single images using micro-texture analysis.
\newblock In: IJCB (2011)

\bibitem{marcel2019handbook}
Marcel, S., Nixon, M.S., Fierrez, J., Evans, N.: Handbook of biometric
  anti-spoofing: Presentation attack detection.
\newblock Springer (2019)

\bibitem{Patel2016Secure}
Patel, K., Han, H., Jain, A.K.: Secure face unlock: Spoof detection on
  smartphones.
\newblock TIFS  (2016)

\bibitem{peixoto2011face}
Peixoto, B., Michelassi, C., Rocha, A.: Face liveness detection under bad
  illumination conditions.
\newblock In: ICIP. IEEE (2011)

\bibitem{pinto2015face}
Pinto, A., Pedrini, H., Schwartz, W.R., Rocha, A.: Face spoofing detection
  through visual codebooks of spectral temporal cubes.
\newblock IEEE TIP  (2015)

\bibitem{qin2021meta}
Qin, Y., Yu, Z., Yan, L., Wang, Z., Zhao, C., Lei, Z.: Meta-teacher for face
  anti-spoofing.
\newblock IEEE TPAMI  (2021)

\bibitem{qin2020learning}
Qin, Y., Zhao, C., Zhu, X., Wang, Z., Yu, Z., Fu, T., Zhou, F., Shi, J., Lei,
  Z.: Learning meta model for zero- and few-shot face anti-spoofing.
\newblock AAAI  (2020)

\bibitem{qureshi2006surveillance}
Qureshi, F.Z., Terzopoulos, D.: Surveillance camera scheduling: A virtual
  vision approach.
\newblock Multimedia systems \textbf{12}(3), 269--283 (2006)

\bibitem{rostami2021detection}
Rostami, M., Spinoulas, L., Hussein, M., Mathai, J., Abd-Almageed, W.:
  Detection and continual learning of novel face presentation attacks.
\newblock In: ICCV (2021)

\bibitem{shao2019multi}
Shao, R., Lan, X., Li, J., Yuen, P.C.: Multi-adversarial discriminative deep
  domain generalization for face presentation attack detection.
\newblock In: CVPR (2019)

\bibitem{shao2019regularized}
Shao, R., Lan, X., Yuen, P.C.: Regularized fine-grained meta face
  anti-spoofing.
\newblock In: AAAI (2020)

\bibitem{siddiqui2016face}
Siddiqui, T.A., Bharadwaj, S., Dhamecha, T.I., Agarwal, A., Vatsa, M., Singh,
  R., Ratha, N.: Face anti-spoofing with multifeature videolet aggregation.
\newblock In: ICPR (2016)

\bibitem{ACER}
international organization~for standardization: Iso/iec jtc 1/sc 37 biometrics:
  Information technology biometric presentation attack detection part 1:
  Framework.
\newblock In: https://www.iso.org/obp/ui/iso (2016)

\bibitem{sun2016context}
Sun, X., Huang, L., Liu, C.: Context based face spoofing detection using active
  near-infrared images.
\newblock In: ICPR (2016)

\bibitem{tan2010face}
Tan, X., Li, Y., Liu, J., Jiang, L.: Face liveness detection from a single
  image with sparse low rank bilinear discriminative model.
\newblock In: ECCV. Springer (2010)

\bibitem{tirunagari2015detection}
Tirunagari, S., Poh, N., Windridge, D., Iorliam, A., Suki, N., Ho, A.T.:
  Detection of face spoofing using visual dynamics.
\newblock IEEE TIFS  (2015)

\bibitem{vareto2020swax}
Vareto, R.H., Saldanha, A.M., Schwartz, W.R.: The swax benchmark: Attacking
  biometric systems with wax figures.
\newblock In: ICASSP (2020)

\bibitem{wang2020cross}
Wang, G., Han, H., Shan, S., Chen, X.: Cross-domain face presentation attack
  detection via multi-domain disentangled representation learning.
\newblock In: CVPR (2020)

\bibitem{wang2021self}
Wang, J., Zhang, J., Bian, Y., Cai, Y., Wang, C., Pu, S.: Self-domain
  adaptation for face anti-spoofing.
\newblock In: AAAI (2021)

\bibitem{wang2009face}
Wang, L., Ding, X., Fang, C.: Face live detection method based on physiological
  motion analysis.
\newblock Tsinghua Science \& Technology \textbf{14}(6), 685--690 (2009)

\bibitem{wang2013face}
Wang, T., Yang, J., Lei, Z., Liao, S., Li, S.Z.: Face liveness detection using
  3d structure recovered from a single camera.
\newblock In: ICB, pp. 1--6. IEEE (2013)

\bibitem{wang2022domain}
Wang, Z., Wang, Z., Yu, Z., Deng, W., Li, J., Li, S., Wang, Z.: Domain
  generalization via shuffled style assembly for face anti-spoofing.
\newblock In: CVPR (2022)

\bibitem{wang2020deep}
Wang, Z., Yu, Z., Zhao, C., Zhu, X., Qin, Y., Zhou, Q., Zhou, F., Lei, Z.: Deep
  spatial gradient and temporal depth learning for face anti-spoofing.
\newblock In: CVPR (2020)

\bibitem{wen2015face}
Wen, D., Han, H., Jain, A.K.: Face spoof detection with image distortion
  analysis.
\newblock IEEE TIFS  (2015)

\bibitem{xu2020improving}
Xu, X., Xiong, Y., Xia, W.: On improving temporal consistency for online face
  liveness detection.
\newblock In: ICCVW (2021)

\bibitem{yang2014learn}
Yang, J., Lei, Z., Li, S.Z.: Learn convolutional neural network for face
  anti-spoofing.
\newblock arXiv preprint arXiv:1408.5601  (2014)

\bibitem{yang2019face}
Yang, X., Luo, W., Bao, L., Gao, Y., Gong, D., Zheng, S., Li, Z., Liu, W.: Face
  anti-spoofing: Model matters, so does data.
\newblock In: CVPR (2019)

\bibitem{yu2022benchmarking}
Yu, Z., Cai, R., Li, Z., Yang, W., Shi, J., Kot, A.C.: Benchmarking joint face
  spoofing and forgery detection with visual and physiological cues.
\newblock arXiv preprint arXiv:2208.05401  (2022)

\bibitem{yu2020face}
Yu, Z., Li, X., Niu, X., Shi, J., Zhao, G.: Face anti-spoofing with human
  material perception.
\newblock In: ECCV (2020)

\bibitem{yu2021revisiting}
Yu, Z., Li, X., Shi, J., Xia, Z., Zhao, G.: Revisiting pixel-wise supervision
  for face anti-spoofing.
\newblock IEEE TBIOM  (2021)

\bibitem{yu2021transrppg}
Yu, Z., Li, X., Wang, P., Zhao, G.: Transrppg: Remote photoplethysmography
  transformer for 3d mask face presentation attack detection.
\newblock IEEE SPL  (2021)

\bibitem{yu2021facial}
Yu, Z., Li, X., Zhao, G.: Facial-video-based physiological signal measurement:
  Recent advances and affective applications.
\newblock IEEE Signal Processing Magazine  (2021)

\bibitem{yu2020multi}
Yu, Z., Qin, Y., Li, X., Wang, Z., Zhao, C., Lei, Z., Zhao, G.: Multi-modal
  face anti-spoofing based on central difference networks.
\newblock In: CVPRW (2020)

\bibitem{yu2021deep}
Yu, Z., Qin, Y., Li, X., Zhao, C., Lei, Z., Zhao, G.: Deep learning for face
  anti-spoofing: A survey.
\newblock IEEE TPAMI  (2022)

\bibitem{yu2021dual}
Yu, Z., Qin, Y., Zhao, H., Li, X., Zhao, G.: Dual-cross central difference
  network for face anti-spoofing.
\newblock In: IJCAI (2021)

\bibitem{yu2020fas2}
Yu, Z., Wan, J., Qin, Y., Li, X., Li, S.Z., Zhao, G.: Nas-fas: Static-dynamic
  central difference network search for face anti-spoofing.
\newblock IEEE TPAMI  (2020)

\bibitem{yu2020searching}
Yu, Z., Zhao, C., Wang, Z., Qin, Y., Su, Z., Li, X., Zhou, F., Zhao, G.:
  Searching central difference convolutional networks for face anti-spoofing.
\newblock In: CVPR (2020)

\bibitem{zhang2021aurora}
Zhang, J., Tai, Y., Yao, T., Meng, J., Ding, S., Wang, C., Li, J., Huang, F.,
  Ji, R.: Aurora guard: Reliable face anti-spoofing via mobile lighting system.
\newblock arXiv preprint arXiv:2102.00713  (2021)

\bibitem{zhang2020face}
Zhang, K.Y., Yao, T., Zhang, J., Tai, Y., Ding, S., Li, J., Huang, F., Song,
  H., Ma, L.: Face anti-spoofing via disentangled representation learning.
\newblock In: ECCV (2020)

\bibitem{zhang2020casia}
Zhang, S., Liu, A., Wan, J., Liang, Y., Guo, G., Escalera, S., Escalante, H.J.,
  Li, S.Z.: Casia-surf: A large-scale multi-modal benchmark for face
  anti-spoofing.
\newblock TBIOM \textbf{2}(2), 182--193 (2020)

\bibitem{casiasurf}
Zhang, S., Wang, X., Liu, A., Zhao, C., Wan, J., Escalera, S., Shi, H., Wang,
  Z., Li, S.Z.: A dataset and benchmark for large-scale multi-modal face
  anti-spoofing.
\newblock In: CVPR (2019)

\bibitem{zhang2020celeba}
Zhang, Y., Yin, Z., Li, Y., Yin, G., Yan, J., Shao, J., Liu, Z.: Celeba-spoof:
  Large-scale face anti-spoofing dataset with rich annotations.
\newblock In: ECCV. Springer (2020)

\bibitem{Zhang2012A}
Zhang, Z., Yan, J., Liu, S., Lei, Z., Yi, D., Li, S.Z.: A face antispoofing
  database with diverse attacks.
\newblock In: ICB (2012)

\end{thebibliography}
}

\end{document}